%% file: main.tex
\documentclass{article}
\pdfoutput=1

\usepackage[nonatbib,final]{neurips_2020}

\usepackage[utf8]{inputenc} 
\usepackage[T1]{fontenc}    
\usepackage{hyperref}       
\usepackage{url}            
\usepackage{booktabs}       
\usepackage{amsfonts}       
\usepackage{amssymb}
\usepackage{nicefrac}       
\usepackage{microtype}      
\usepackage{bbm}
\usepackage{tikz}
\usetikzlibrary{arrows, calc}
\usepackage{hyperref}
\usepackage{layouts}

\usepackage{graphicx}
\usepackage{subfigure}
\usepackage{mathtools}
\usepackage{bm}
\usepackage[]{algorithm2e}
\usepackage{amsthm}

\usepackage{pgfplots}
\pgfplotsset{compat=1.14}
\pgfplotsset{
    /pgf/number format/res/.style={
        fixed,
        fixed zerofill,
        precision=2,
        skip 0.,
    },
}
\pgfplotsset{
    /pgf/number format/perc/.style={
        fixed,
        precision=0,
    },
}
\usepackage{pgf}
\usepackage{datatool}
\DTLsetseparator{,}

\DeclareMathOperator*{\argmax}{argmax}

\DeclareMathOperator*{\ch}{ch}
\DeclareMathOperator*{\pa}{pa}
\DeclareMathOperator*{\scope}{sc}

\newcommand{\cbar}{\,|\,}
\newcommand{\X}{\mathbf{X}}
\newcommand{\x}{\mathbf{x}}
\newcommand{\xs}{\mathcal{X}}
\newcommand{\xxs}{\bm{\mathcal{X}}}
\newcommand{\ys}{\mathcal{Y}}
\newcommand{\data}{\mathcal{D}}
\newcommand{\graph}{\mathcal{G}}
\newcommand{\lcellsfamily}{\bm{\mathcal{F}}}
\newcommand{\lcells}{\bm{\mathcal{A}}}
\newcommand{\lcell}{\ensuremath{\mathcal{A}}}
\newcommand{\leaf}{l}
\newcommand{\indf}[1]{\ensuremath{{\mathbbm{1}(#1)}}}
\newcommand{\indfu}[1]{\ensuremath{{\mathbbm{1}_{#1}}}}
\newcommand{\node}{v}
\newcommand{\nodec}{u}

\newcommand{\productnode}{\pi}
\newcommand{\w}{w}
\newcommand{\prd}{\ensuremath{\mathbb{P}}}
\newcommand{\pat}{\ensuremath{\Lambda}}
\newcommand{\ntrees}{\ensuremath{n_t}}
\newcommand{\asymO}{{\ensuremath{\mathcal{O}}}}
\newcommand{\noto}{{\ensuremath{\lnot o}}}
\newcommand{\GeDT}{GeDT}
\newcommand{\GeF}{GeF}
\newcommand{\Gp}{GeF$^+$}

\newcommand{\mycap}{Accuracy against proportion of missing values. The same plot is repeated ten times, each time highlighting a different method and its 95\% confidence interval.}

\newtheorem*{rep@theorem}{\rep@title}
\newcommand{\newreptheorem}[2]{%
\newenvironment{rep#1}[1]{%
 \def\rep@title{#2 \ref{##1}}%
 \begin{rep@theorem}}%
 {\end{rep@theorem}}}
\makeatother

\newreptheorem{lemma}{Lemma}
\newtheorem{theorem}{Theorem}
\newreptheorem{theorem}{Theorem}
\newtheorem{proposition}{Proposition}
\newtheorem{corollary}{Corollary}
\newreptheorem{corollary}{Corollary}
\newtheorem{definition}{Definition}

\newtheoremstyle{named}{}{}{\itshape}{}{\bfseries}{.}{.5em}{#1 \thmnote{\textbf{#3}}}
\theoremstyle{named}

\newcommand{\maketable}[2]{
\DTLloaddb[]{data}{#1}
\DTLforeach{data}{\l=lower}
{
    \def\theMax{0}                                  
    \DTLforeachkeyinrow{\thisValue}{                
        \ifthenelse{\dtlcol>1}{                     
            \DTLmax{\theMax}{\theMax}{\thisValue}   
        }{}
    } 

    \DTLforeachkeyinrow{\thisValue}{                
        \ifthenelse{\dtlcol>1}{                     
            \ifthenelse{\DTLisFPgteq{\thisValue}{\l}}{
                \DTLreplaceentryforrow{\dtlkey}{\textbf{\thisValue}}
            }{}
            \ifthenelse{\DTLisieq{\thisValue}{\theMax}}{
                \DTLreplaceentryforrow{\dtlkey}{\underline{\textbf{\theMax}}}
            }{}
        }{}
    }     
}
\begin{table}[h!]
\fontsize{9}{9}\selectfont
\setlength{\tabcolsep}{2.25pt}
\begin{center}
    \caption{Accuracy per percent of missing values at test time with 95\% confidence intervals.}
    \label{#2}
    \begin{tabular}{c | c c c c c}
    \toprule (\%) & Surrogate & Friedman & Mean & KNN & MissForest
    \DTLforeach*{data}{\x=Missing, \surr=Surr, \sci=Sci, 
                        \fried=Friedman, \fci=Fci,
                        \mean=Mean, \mci=Mci,
                        \knn=KNN, \kci=Kci,
                        \mf=MissForest, \mfci=MFci,
                        \gef=GeF, \gefci=GeFci,
                        \gefl=GeF(LSPN), \geflci=GeF(LSPN)ci,
                        \gefp=GeFp, \gefpci=GeFpci,
                        \gefpl=GeFp(LSPN), \gefplci=GeFp(LSPN)ci}
    {
        \DTLiffirstrow{\\ \midrule}{\\}
        \pgfmathprintnumber[perc]\x &
        \surr~$\pm$~\pgfmathprintnumber[res]\sci &
        \fried~$\pm$~\pgfmathprintnumber[res]\fci &
        \mean~$\pm$~\pgfmathprintnumber[res]\mci &
        \knn~$\pm$~\pgfmathprintnumber[res]\kci &
        \mf~$\pm$~\pgfmathprintnumber[res]\mfci
    }
    \\\bottomrule
    \end{tabular}
    
    \vspace{6pt}
    
    \begin{tabular}{c | c c c c c }
    \toprule (\%) & LearnSPN & GeF & GeF(LSPN) & $\text{GeF}^+$ & $\text{GeF}^+$(LSPN)
    \DTLforeach*{data}{\x=Missing, \surr=Surr, \sci=Sci, 
                        \fried=Friedman, \fci=Fci,
                        \mean=Mean, \mci=Mci,
                        \knn=KNN, \kci=Kci,
                        \mf=MissForest, \mfci=MFci,
                        \gef=GeF, \gefci=GeFci,
                        \gefl=GeF(LSPN), \geflci=GeF(LSPN)ci,
                        \gefp=GeFp, \gefpci=GeFpci,
                        \gefpl=GeFp(LSPN), \gefplci=GeFp(LSPN)ci,
                        \lspn=LSPN, \lspnci=LSPNci}
    {
        \DTLiffirstrow{\\ \midrule}{\\}
        \pgfmathprintnumber[perc]\x &
        \lspn~$\pm$~\pgfmathprintnumber[res]\lspnci &
        \gef~$\pm$~\pgfmathprintnumber[res]\gefci &
        \gefl~$\pm$~\pgfmathprintnumber[res]\geflci &
        \gefp~$\pm$~\pgfmathprintnumber[res]\gefpci &
        \gefpl~$\pm$~\pgfmathprintnumber[res]\gefplci 
    }
    \\\bottomrule
    \end{tabular}   
    
\end{center}
\end{table}
\DTLgdeletedb{data}
}

\usepackage{times}
\usepackage[title]{appendix}

\title{Joints in Random Forests}
          
\author{ {\bf Alvaro H. C. Correia} \\ a.h.chaim.correia@tue.nl \\ Eindhoven University of Technology \\ 
\And {\bf Robert Peharz}  \\ r.peharz@tue.nl \\ Eindhoven University of Technology \\
\And {\bf Cassio de Campos} \\ c.decampos@tue.nl \\ Eindhoven University of Technology \\
}

\begin{document}

\maketitle

\begin{abstract}
Decision Trees (DTs) and Random Forests (RFs) are powerful discriminative learners and tools of central importance to the everyday machine learning practitioner and data scientist. Due to their discriminative nature, however, they lack principled methods to process inputs with missing features or to detect outliers, which requires pairing them with imputation techniques or a separate generative model. In this paper, we demonstrate that DTs and RFs can naturally be interpreted as generative models, by drawing a connection to Probabilistic Circuits, a prominent class of tractable probabilistic models. This reinterpretation equips them with a full joint distribution over the feature space and leads to Generative Decision Trees (GeDTs) and Generative Forests (GeFs), a family of novel hybrid generative-discriminative models. This family of models retains the overall characteristics of DTs and RFs while additionally being able to handle missing features by means of marginalisation. Under certain assumptions, frequently made for Bayes consistency results, we show that consistency in GeDTs and GeFs extend to any pattern of missing input features, if missing at random. Empirically, we show that our models often outperform common routines to treat missing data, such as K-nearest neighbour imputation, and moreover, that our models can naturally detect outliers by monitoring the marginal probability of input features.
\end{abstract}

\section{Introduction}

Decision Trees (DTs) and Random Forests (RFs) are probably the most widely used non-linear machine learning models of today. While Deep Neural Networks are in the lead for image, video, audio, and text data---likely due to their beneficial inductive bias for signal-like data---DTs and RFs are, by and large, the default predictive model for tabular, domain-agnostic datasets. Indeed, Kaggle's 2019 report on the \emph{State of Data Science and Machine Learning} \cite{Kaggle2019} lists DTs and RFs as second most widely used techniques, right after linear and logistic regressions. Moreover, a study by Fernandez et al.~\cite{Fernandez2014} found that RFs performed best on 121 UCI datasets against 179 other classifiers. Thus, it is clear that DTs and RFs are of central importance for the current machine learning practitioner.

DTs and RFs are generally understood as \emph{discriminative models}, that is, they are solely interpreted as predictive models, such as \emph{classifiers} or \emph{regression functions}, while attempts to additionally interpret them as \emph{generative} models are scarce. In a nutshell, the difference between discriminative and generative models is that the former aim to capture the \emph{conditional distribution} $P(Y \cbar \X)$, while the latter aim to capture the whole \emph{joint distribution} $P(Y,\X)$, where $\X$ are the input features and $Y$ is the variable to be predicted---discrete for classification and continuous for regression. In this paper, we focus on classification, but the extension to regression is straightforward.

Generative and discriminative models are rather complementary in their strengths and use cases.
While discriminative models typically fare better in predictive performance, generative models allow to analyse and capture the structure present in the input space. They are also ``all-round predictors'', that is, not restricted to a single prediction task but also capable of predicting any $X$ given $Y \cup \X \setminus X$. Moreover, generative models have some crucial advantages on the prediction task $P(Y \cbar \X)$ a discriminative model has been trained on, as they naturally allow to \emph{detect outliers} (by monitoring $P(\X)$) and \emph{treat missing features} (by marginalisation). A purely discriminative model does not have any ``innate'' mechanisms to deal with these problems, and needs to be supported with a generative model $P(\X)$ (to detect outliers) or imputation techniques (to handle missing features).

Ideally, we would like the best of both worlds: having the good predictive performance of discriminative models \emph{and} the advantages of generative models. In this paper, we show that this is achievable for DTs and RFs by relating them to \emph{Probabilistic Circuits} (PCs) \cite{VanDenBroeck2019}, a class of generative models based on computational graphs of \emph{sum nodes} (mixtures), \emph{product nodes} (factorisations), and \emph{leaf nodes} (distribution functions). PCs subsume and represent a wide family of related models, such as \emph{arithmetic circuits} \cite{Darwiche2003}, \emph{AND/OR-graphs} \cite{Marinescu2005}, \emph{sum-product networks} \cite{Poon2011}, \emph{cutset networks} (CNets) \cite{Rahman2014}, and \emph{probabilistic sentential decision diagrams} \cite{Kisa2014}. While many researchers are aware of the similarity between DTs and PCs---most notably, CNets \cite{Rahman2014} can be seen as a type of generative DT---the connection to classical, discriminative DTs \cite{Quinlan1986} and RFs \cite{Breiman2001} has not been studied so far.

We show that DTs and RFs can be naturally cast into the PC framework. For any given DT, we can construct a corresponding PC, a \emph{Generative Decision Tree} (\GeDT{}), representing a full joint distribution $P(Y,\X)$. This distribution gives rise to the predictor $P(Y \cbar \X) = \nicefrac{P(Y, \X)}{\sum_y P(y, \X)}$, which is identical to the original DT, if we impose certain constraints on the conversion from DT to \GeDT{}. Additionally, a \GeDT{} also fits the joint distribution $P(\X)$ to the training data, ``upgrading'' the DT to a fully generative model. For a \emph{completely observed sample} $\X = \x$, the original DT and a corresponding \GeDT{} agree entirely (yield the exact same predictions), and moreover, have the \emph{same computational complexity} (a discussion on time complexity is deferred to Appendix B). By converting each DT in an RF into an \GeDT{}, we obtain an ensemble of \GeDT{}s, which we call \emph{Generative Forest} (\GeF{}). Clearly, if each \GeDT{} in a \GeF{} agrees with its original DT, then \GeF{}s also agree with their corresponding RFs.

\GeDT{}s and \GeF{}s have a crucial advantage in the case of \emph{missing features}, that is, assignments $\X_o = \x_o$ for some subset $\X_o \subset \X$, while $\X_{\noto} = \X \setminus \X_o$ are \emph{missing at random}.
In a \GeDT{}, we can marginalise the missing features and yield the predictor 
\begin{equation}   \label{eq:marginalized_classifier}
P(Y \cbar \X_o) = 
\frac{\int_{\x_{\noto}} P(Y, \X_o, \x_{\noto})\mathrm{d}\x_{\noto}}
{\sum_y \int_{\x_{\noto}} P(y, \X_o, \x_{\noto}) \mathrm{d}\x_{\noto}}.
\end{equation}
For \GeF{}s, we yield a corresponding ensemble predictor for missing features, by applying marginalisation to each \GeDT{}. Using the true data generating distribution in Eq.~\eqref{eq:marginalized_classifier} would deliver the \emph{Bayes optimal predictor} for any subset $\X_o$ of observed features. Thus, since \GeDT{}s are trained to approximate the true distribution, using the predictor of Eq.~\eqref{eq:marginalized_classifier} under missing data is well justified. We show \GeDT{}s are in fact \emph{consistent}: they converge to the Bayes optimal classifier as the number of data points goes to infinity. Our proof requires similar assumptions to those of previous results for DTs \cite{Biau2008,Breiman1984,Gordon1978} but is substantially more general: while consistency in DTs is shown only for a classifier $P(Y \cbar \X)$ using fully observed samples, our consistency result holds for \emph{all} $2^{|X|}$ classifiers $P(Y \cbar \X_o)$: one for each observation pattern $\X_o \subseteq \X$. While the high-dimensional integrals in Eq.~\eqref{eq:marginalized_classifier} seem prohibitive, they are in fact \emph{tractable}, since a remarkable feature of PCs is that computing any marginal has \emph{the same complexity} as evaluating the full joint, namely linear in the \emph{circuit size}.

This ability of our models is desirable, as there is no clear consensus on how to deal with missing features in DTs at test time: The most common strategy is to use \emph{imputation}, e.g.~\emph{mean} or \emph{k-nearest-neighbour (KNN) imputation}, and subsequently feed the completed sample to the classifier. DTs also have two ``built-in'' methods to deal with missing features that do not require external models. These are the so-called \emph{surrogate splits} \cite{Therneau1997} and an unnamed method proposed by Friedman in 1977 \cite{Friedman1977,Quinlan1987b}. Among these, KNN imputation seems to be the most widely used, and typically delivers good results on real-world data. However, we demonstrate it does not lead to a consistent predictor under missing data, even when assuming idealised settings. Moreover, in our experiments, we show that \GeF{} classification under missing inputs often outperforms standard RFs with KNN imputation.

Our generative interpretation can be easily incorporated in existing DT learners and does not require drastic changes in the learning and application practice for DTs and RFs. Essentially, any DT algorithm can be used to learn \GeDT{}s, requiring only minor bookkeeping and some extra generative learning steps. There are de facto no model restrictions concerning the additional generative learning steps, representing a generic scheme to augment DTs and RFs to generative models.

\section{Notation and Background}

In this paper we focus on classification tasks.
To this end, let the set of explanatory variables (features) be $\X = \{X_1, X_2, \ldots, X_m\}$, where continuous $X_i$ assume values in some compact set $\xs_i \subset \mathbb{R}$ and discrete $X_i$ assume values in $\xs_i = \{1, \ldots, K_i\}$, where $K_i$ is the number of states for $X_i$. Let the joint \emph{feature space} of $\X$ be denoted as $\xxs$. We denote joint states, i.e.~elements from $\xxs$, as $\x$ and let $\x[i]$ be the state in $\x$ belonging to $X_i$. The class variable is denoted as $Y$, assuming values in $\ys =\{1, \ldots, K\}$, where $K$ is the number of classes.    We assume that the pair $(\X,Y)$ is drawn from a fixed joint distribution $\mathbb{P}^*(\X, Y)$ which has density $p^*(\X, Y)$. While the true distribution $\mathbb{P}^*$ is unknown, we assume that we have a dataset $\data_n = \{(\x_1,y_1), \ldots, (\x_n,y_n)\}$ of $n$ i.i.d.~samples from $\mathbb{P}^*$. When describing a directed graph $\graph$, we refer to its set of nodes as $V$, reserving letters $u$ and $v$ for individual nodes. We denote the set of children and parents of a node $v$ as $\ch(v)$ and $\pa(v)$, respectively. Nodes $v$ without children are referred to as \emph{leaves}, and nodes without parents are referred to as \emph{roots}.

\textbf{Decision Trees.}
A \emph{decision tree} (DT) is based on a \emph{rooted directed tree} $\graph$, i.e.~an acyclic directed graph with exactly one root $v_r$ and whose other nodes have exactly one parent. Each node $v$ in the DT is associated with a \emph{cell} $\xxs_v$, which is a subset of the feature space $\xxs$. The cell of the root node $v_r$ is the whole $\xxs$. The \emph{child cells} of node $v$ form a partition of $\xxs_v$, i.e.~ $\bigcup_{u \in \ch(v)} \xxs_u = \xxs_v$, $\xxs_u \cap \xxs_{u'} = \emptyset, ~\forall u,u' \in \ch(v)$. These partitions are usually defined via  \emph{axis-aligned splits}, by associating a decision variable $X_i$ to $v$, and partitioning the cell according to some rule on $X_i$'s values. Formally, we first project $\xxs_v$ onto its $i$\textsuperscript{th} coordinate, yielding $\xs_{i,v} := \{\x[i] ~|~ \x \in \xxs_v \}$, and construct a partition $\{\xs_{i,u}\}_{u \in \ch(v)}$ of $\xs_{i,v}$. The child cells are then given by $\xxs_u = \{ \x ~|~ \x \in \xxs_v \land \x[i] \in \xs_{i,u} \}$. Common choices for this partition are \emph{full splits} for discrete variables, i.e.~choosing $\{\xs_{i,u}\}_{u \in \ch(v)} = \{\{x_i\}\}_{x_i \in \xs_{i,v}}$ where children $u$ and states $x_i$ are in one-to-one correspondence, and \emph{thresholding} for continuous variables, i.e.~choosing $\{\xs_{i,u}\}_{u \in \ch(v)} = \{\{x_i < t\}, \{x_i \geq t\}\}$ for some threshold $t$. Note that the leaf cells of a DT represent a partition $\lcells$ of the feature space $\xxs$. We denote the elements of $\lcells$ as $\lcell$ and define $\lcell_v = \xxs_v$ for each leaf $v$. A \emph{DT classifier} is constructed by equipping each $\lcell \in \lcells$ with a classifier $f^{\lcell} \colon \lcell \mapsto \Delta^K$, where $\Delta^K$ is the set of probability distributions over $K$ classes, i.e.~$f^{\lcell}$ is a conditional distribution defined on $\lcell$. This distribution is typically stored as absolute class counts of the training samples contained in $\lcell$.

The overall DT classifier is given as $f(\x) = f^{\lcell(\x)}(\x)$ where $\lcell(\x)$ is the leaf cell containing $\x$; $\lcell(\x)$ is found by parsing the DT top-down, following the partitions (decisions) consistent with $\x$.
This formulation captures the vast majority of DT classifiers proposed in the literature, notably CART~\cite{Breiman1984} and ID3~\cite{Quinlan1986}. The probably most widely used variant of DTs---which we also assume in this paper---is to define $f^\lcell$ as a constant function, returning the class proportions in cell $\lcell$. The $\arg\max$ of $f^\lcell(\x)$ is equivalent to majority voting among all training samples which fall into the same cell. When learning a DT, the number of available training samples per cell reduces quickly, which leads to overfitting and justifies the need for pruning techniques \cite{Breiman1984,Mingers1987,Quinlan1986,Quinlan1987}.

\textbf{Random Forests.} \emph{Random Forests} (RFs) are ensembles of DTs which effectively counteract overfitting.
Each DT in a RF is learned in a randomised fashion by, at each learning step, drawing a random sub-selection of variables containing only a fraction $p$ of all variables, where typical values are $p=0.3$ or $p=\sqrt{m}$. The resulting DTs are not pruned but made ``deep'' until each leaf cell contains either only samples of one class or less than $T$ samples, where typical values are $T \in \{1,5,10\}$. This yields low bias, but high variance in the randomised DTs, which makes them good candidates for \emph{bagging} (bootstrap aggregation) \cite{Hastie2009}. Thus, to further increase the variability among the trees, each of them is learned on a \emph{bootstrapped} version of the training data \cite{Breiman2001}.

\textbf{Probabilistic Circuits.}
In this paper, we relate DTs to \emph{Probabilistic Circuits} (PCs) \cite{VanDenBroeck2019}, a family of density representations facilitating many exact and efficient inference routines. PCs are, like DTs, based on a rooted acyclic directed graph $\graph$, albeit one with different semantics. PCs are computational graphs with three types of nodes, namely i) distribution nodes, ii) sum nodes and iii) product nodes. Distribution nodes are the leaves of the graph $\graph$, while sum and product nodes are the internal nodes. Each distribution node (leaf) $\node$ computes a probability density\footnote{By an adequate choice of the underlying measure, this also subsumes probability mass functions.} over some subset $\X' \subseteq \X$, i.e.~a normalised function $p_\node(\x') \colon \xxs' \mapsto \mathbb{R}^{+}$ from the state space of $\X'$ to the non-negative real numbers. The set of variables $\X'$ over which the leaf computes a distribution is called the \emph{scope} of $v$, and denoted by $\scope(\node) := \X'$. Given the scopes of the leaves, the scope of any internal node $\node$ (sum or product) is recursively defined as $\scope(\node) = \cup_{\nodec \in \ch(\node)} \scope(\nodec)$. Sum nodes compute convex combinations over their children, i.e.~if $\node$ is a sum node, then $\node$ computes $\node(\x) = \sum_{\nodec \in \ch(\node)} \w_{\node,\nodec} \nodec(\x)$, where $\w_{\node,\nodec} \geq 0$ and $\sum_{\nodec \in \ch(\node)} \w_{\node,\nodec} = 1$. Product nodes compute the product over their children, i.e.~if $\node$ is a product node, then $\node(\x) = \prod_{\nodec \in \ch(\node)} \nodec(\x)$. The density $p(\X)$ represented by a PC is the function computed by its root node, and can be evaluated with a feedforward pass.

The main feature of PCs is that they facilitate a wide range of \emph{tractable} inference routines, which go hand in hand with certain structural properties, defined as follows \cite{Darwiche2003, VanDenBroeck2019}: i) A sum node $\node$ is called \emph{smooth} if its children have all the same scope: $\scope(\nodec) = \scope(\nodec')$, for any $\nodec, \nodec' \in \ch(\node)$. ii) A product node $\node$ is called \emph{decomposable} if its children have non-overlapping scopes: $\scope(\nodec) \cap \scope(\nodec') = \emptyset$, for any $\nodec, \nodec' \in \ch(\productnode)$, $\nodec \not= \nodec'$. A PC is smooth (respectively decomposable) if all its sums (respectively products) are smooth (respectively decomposable). Smoothness and decomposability are sufficient to ensure \emph{tractable marginalisation} in PCs. In particular, assume that we wish to evaluate the density over $\X_o \subset \X$ for evidence $\X_o = \x_o$, while marginalising $\X_\noto = \X \setminus \X_o$. In PCs, this task reduces to performing marginalisation at the leaves \cite{Peharz2015}, that is, for each leaf $\node$ one marginalises $\scope(\node) \cap \X_\noto$, and evaluates it for the values corresponding to $\scope(\node) \cap \X_o$. The desired marginal $p_{\X_o}(\x_o)$ results from evaluating internal nodes as in computing the complete density. Furthermore, a PC is called \emph{deterministic} \cite{Darwiche2003, VanDenBroeck2019} if it holds that for each complete sample $\x$, each sum node has at most one non-zero child.Determinism and decomposability are sufficient conditions for \emph{efficient maximisation}, which again, like density evaluation and marginalisation, reduces to a single feedforward pass. 

\section{Generative Decision Trees}

Given a learned DT and the dataset $\data = \{(\x_1,y_1), \dots, (\x_n,y_n)\}$ it has been learned on, we can obtain a corresponding generative model, by converting the DT into a PC.
This conversion is given in Algorithm~\ref{algo:DT2PC}.
\begin{algorithm}[t]
\SetAlgoLined
\SetKwInOut{Input}{Input}
\SetKwInOut{Output}{Output}
\Input{Decision Tree $\graph$ and training data $\data$}
\Output{Probabilistic Circuit $\graph'$}
let $\graph'$ be a structural copy of $\graph$ and
let $v'$ be the node in $\graph'$ which corresponds to $v$ of $\graph$\\
for root node $v$ of $\graph$, set $\data_v = \data$\\
\For{$v$ \textbf{\emph{in}} topdownsort($V$)}{
\eIf{$v$ is internal}
  {
  get partition $\xs_{i,u}$ of decision variable $X_i$ associated with $v$ \\
  \For{$u \in \ch(v)$}{
    let $w_{v'u'} = \frac{\sum_{\x \in \data_v} \indf{\x[i] \in \xs_{i,u}}}{|\data_v|}$ \\
    set $\data_u = \{\x \in \data_v ~|~ \x[i] \in \xs_{i,u}\}$
  }
  let $v'$ be a sum node $\sum_{u' \in \ch(v')} w_{v'u'} u'$
  }
  {
  let $v'$ be a density $p_{v'}(\x,y)$ with support $\lcell_v$, learned from $\data_{v}$}
  }
\caption{Converting DT to PC (\GeDT{}).}
\label{algo:DT2PC}
\end{algorithm}
In a nutshell, Algorithm~\ref{algo:DT2PC} converts each decision node into a sum node and each leaf into a density with support restricted to the leaf's cell. The training samples can be figured to be routed from the root node to the leaves, following the decisions at each decision/sum node. The sum weights are given by the fraction of samples which are routed from the sum node to each of its children. The leaf densities are learned on the data which arrives at the respective leaves.

 \begin{figure*}
     \centering
     \input{tree1.tex}
     \caption{Illustration of a DT and its corresponding PC as obtained by Algorithm~\ref{algo:DT2PC}.}
     \label{fig:dt-spn}
 \end{figure*}
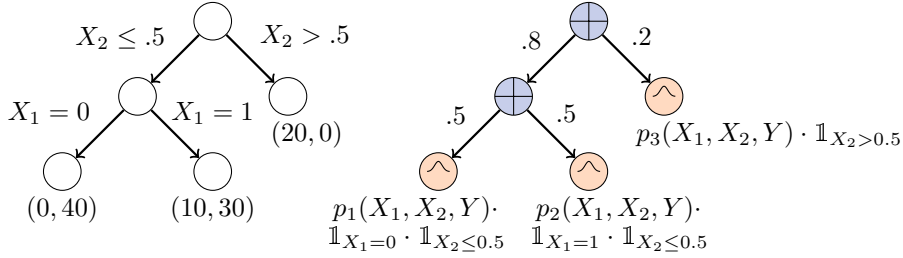

As an example, assuming $\X$ and $Y$ factorise at the leaves, Algorithm~\ref{algo:DT2PC} applied to the DT on the left-hand side of Figure~\ref{fig:dt-spn} gives the PC on the right-hand side and these densities at the leaves:
\begin{align*}
p_1(X_1,X_2,Y) &= p_1(X_1,X_2)  (0\cdot \indf{Y=0} + 1\cdot \indf{Y=1}),\\
p_2(X_1,X_2,Y) &= p_2(X_1,X_2) (0.25\cdot \indf{Y=0} + 0.75\cdot \indf{Y=1}),\\
p_3(X_1,X_2,Y) &= p_3(X_1,X_2) (1\cdot \indf{Y=0} + 0\cdot \indf{Y=1})\, ,
\end{align*}
\noindent 
Note that $X_1$ is deterministic (all mass absorbed in one state) in  $p_1$ and $p_2$, since $X_1$ has been fixed by the tree construction, while $p_3$ is a ``proper'' distribution over $X_1$ and $X_2$. Densities $p_i(X_1,X_2)$ do not appear in the DT representation and illustrate the extension brought in by the PC formalism.

We denote the output of Algorithm~\ref{algo:DT2PC} as a \emph{Generative Decision Tree} (\GeDT{}). Note that \GeDT{}s are proper PCs over $(\X,Y)$, albeit rather simple ones: they are tree-shaped and contain only sum nodes. They are clearly smooth, since each leaf density has the full scope $(\X,Y)$, and they are trivially decomposable, as they do not contain products.
Thus, both the full density or any sub-marginal can be evaluated by simply evaluating the \GeDT{} bottom up, where for marginalisation tasks we first need to perform marginalisation at the leaves. Furthermore, it is easy to show that any \GeDT{} is \emph{deterministic} (see Appendix A). As shown in \cite{Peharz2014,Rahman2014}, the sum-weights set by Algorithm~\ref{algo:DT2PC} are in fact the \emph{maximum likelihood} weights for deterministic PCs.

In Algorithm~\ref{algo:DT2PC}, we learn a density $p_{\node}(\x, y)$ for each leaf $\node$, where we have not yet specified the model or learning algorithm. Thus, we denote \GeDT{}($M$) as a \GeDT{} whose leaf densities are learned by ``method $M$'', where $M$ might be graphical models, again PCs, or even neural-based density estimators \cite{Kingma2014, Rezende2015}. In order to ensure tractable marginalisation of the overall \GeDT{}, however, we use either \emph{fully factorised} leaves---for each leaf $v$, $p_{v}(\X, Y) =  p_{v}(X_1)p_{v}(X_2) \ldots p_{v}(X_m)p_{v}(Y)$---or PCs learned with \emph{LearnSPN} \cite{Gens2013}. In both cases marginalisation at the leaves, and hence in the whole \GeDT{}, is efficient. Regardless of the model $M$, we generally learn the leaves using the maximum likelihood principle, or some proxy of it. Thus, since the sum-weights are already set to the (global) maximum likelihood solution by Algorithm~\ref{algo:DT2PC}, the overall \GeDT{} also fits the training data. A basic design choice is how to model the dependency between $\X$ and $Y$ at the leaves: we might assume independence between them, i.e.~assume $p(\X, Y) = p(\X) p(Y)$ (\emph{class-factorised} leaves)\footnote{Note that such independence is only a context-specific one, conditional on the state of variables associated with sum nodes \cite{Peharz2016,Poon2011}. This assumption does not represent global independence between $\X$ and $Y$.} or simply pass the data over both $\X$ and $Y$ to a learning algorithm and let it determine the dependency structure itself (\emph{full} leaves). Note that we are free to have different types of density estimators for different leaves in a single \GeDT. A natural design choice is to match the complexity of the estimator in a leaf to the number of samples it contains.

The main semantic difference between DTs and \GeDT{}s is that a DT represents a classifier, i.e.~a conditional distribution $f(\x)$, while the corresponding \GeDT{} represents a full joint distribution $p(\X, Y)$. The latter naturally lends itself towards classification by deriving the conditional distribution $p(Y \cbar \x) \propto p(\x, Y)$. How are the original DT classifier $f(\x)$ and the \GeDT{} classifier $p(Y \cbar \x)$ related? In theory, $p(Y \cbar \x)$ might differ substantially from $f(\x)$, since every feature might influence classification in a \GeDT{}, even if it never appears in any decision node of the DT. In the case of class-factorised leaves, however, we obtain ``backwards compatibility''.
\begin{theorem}   \label{theo:backwards_compatible}
Let $f$ be a DT classifier and $p(Y \cbar \x)$ be a corresponding \GeDT{} classifier, where each leaf in \GeDT{} is class-factorised, i.e.~of the form $p(Y)p(\X)$, and where $p(Y)$ has been estimated in the maximum-likelihood sense. Then $f(\x) = p(Y \cbar \x)$, provided that $p(\x) > 0$.
\end{theorem}

For space reasons, proofs and complexity results are deferred to the appendix. Theorem~\ref{theo:backwards_compatible} shows that DTs and \GeDT{}s yield exactly the same classifier for class-factorised leaves and complete data. DTs achieve their most impressive performance when used as an ensemble in RFs. It is straight-forward to convert each DT in an RF using Algorithm~\ref{algo:DT2PC}, yielding an ensemble of \GeDT{}s. We call such an ensemble a \emph{Generative Forest} (\GeF{}). This result extends to ensembles, as clearly when all \GeDT{}s in a \GeF{} use class-factorised leaves, then according to Theorem~\ref{theo:backwards_compatible}, \GeF{}s yield exactly the same prediction function as their corresponding RFs. This means that the everyday practitioner can safely replace RFs with class-factorised \GeF{}s, gaining the ability to classify under \emph{missing input data}.

\section{Handling Missing Values}

The probably most frequent strategy to treat missing inputs in DTs and RFs is to use some \emph{single imputation} technique, i.e.~to first predict any missing values based on the observed ones, and then use the imputed sample as input to the classifier. A particularly prominent method is \emph{K-nearest neighbour} (KNN) imputation, which typically works well in practice. This strategy, however, is not Bayes consistent and can in principle be arbitrarily bad. This can be shown with a simple example. Assume two multivariate Gaussian features $X_1$ and $X_2$ with $var(X_1) \geq \tau$, $var(X_2) \geq \tau$ for some $\tau > 0$, i.e.~the variances of $X_1$ and $X_2$ are bounded from below. Let the conditional class distribution be $p(y \cbar x_1, x_2) = \indf{|x_2 - \mathbb{E}[X_2 \cbar x_1]| > \epsilon}$, i.e.~$Y$ detects whether $X_2$ deviates more than $\epsilon$ from its mean, conditional on $X_1$. Assume $X_2$ is missing and use KNN to impute it, based on $X_1=x_1$. KNN is known to be a consistent regressor, provided the number of neighbours goes to infinity but vanishes in comparison to the number of samples \cite{Devroye1996}. Thus, the imputation for $X_2$ based on $x_1$ converges to $\mathbb{E}[X_2 \cbar x_1],$ yielding a constant prediction of $Y=0$. It follows that by making $\epsilon$ arbitrarily small, we can push the classification error arbitrarily close to $1$, while the true error goes to $0$.

Assuming that inputs are missing at random \cite{Little2019} and that we have only inputs $\x_o$ for some subset $\X_o \subset \X$, a \GeDT{} naturally yields a classifier $p(y \cbar \X_o)$, by marginalising missing features as in Eq.~\eqref{eq:marginalized_classifier}. Recall that marginalisation in PCs, and thus in \GeDT{}s, can be performed with a single feedforward pass, given that the \GeDT{}'s leaves permit efficient marginalisation. In our experiments, we use either fully factorised leaves or PC leaves learned by LearnSPN \cite{Gens2013}, a prominent PC learner, such that we can efficiently and exactly evaluate $p(Y \cbar \X_o)$ with a single pass through the network. Thus, a \GeDT{} represents in fact $2^{|\X|}$ classifiers, one for each missingness pattern. Since the true data distribution yields Bayes optimal classifiers for each $\X_o$, and since the parameters of \GeDT{}s are learned in the maximum likelihood sense, using the \GeDT{} predictor $p(y \cbar \X_o)$ for missing data is natural. For a simplified variant of \GeDT{}s, we can show that they converge to the true distribution and are therefore Bayes consistent classifiers for each $\X_o$. Theorem~\ref{theo:consistency} assumes, without loss of generality, that all variables in $\X$ are continuous.
\begin{theorem} \label{theo:consistency}
Let $\mathbb{P}^*$ be an unknown data generating distribution with density $p^*(\X,Y)$, and let $\mathcal{D}_n$ be a dataset drawn i.i.d.~from $\mathbb{P}^*$. Let $\graph$ be a DT learned with a DT learning algorithm, using axis-aligned splits. Let $\lcells^n$ be the (rectangular) leaf cells produced by the learning algorithm. Assume it holds that i) $\lim_{n \rightarrow \infty}\nicefrac{|\lcells^n| \log(n)}{n} \rightarrow 0$ and ii) $\mathbb{P}^*(\{\x ~|~ \mathrm{diam}(\lcell^n_x) > \gamma\}) \rightarrow 0$ almost surely for all $\gamma > 0$, where $\mathrm{diam}(\lcell)$ is the diameter of cell $\lcell$. Let $\graph'$ be the \GeDT{} corresponding to $\graph$, obtained via Algorithm~\ref{algo:DT2PC}, where for each leaf $v$, $p_v$ is of the form $p_v(Y) p_v(\X)$, with $p_v(\X)$ uniform on $\lcell_v$ and $p_v(Y)$ the maximum likelihood Categorical (fractions of class values of samples in $\lcell_v$). Then the \GeDT{} distribution is $l_1$-consistent, i.e.~$\sum_y \int | p(\x,y) - p^*(\x,y) | \mathrm{d}\x \rightarrow 0$, almost surely.
\end{theorem}

Note that the assumptions in Theorem~\ref{theo:consistency} are in line with consistency results for DTs. See for example \cite{Breiman1984,Devroye1996,Lugosi1996}, all of which require, in some sense, that the number of cells vanishes in comparison to the number of samples, and that the cell sizes shrink to zero. Theorem~\ref{theo:consistency} naturally leads to the Bayes-consistency of \GeDT{}s and \GeF{}s under missing inputs.

\begin{corollary} \label{cor:gedt_consistency}
Under assumptions of Theorem~\ref{theo:consistency}, any \GeDT{} predictor $p(Y \cbar \X_o)$, for $\X_o \subseteq \X$ is Bayes consistent.
\end{corollary}
\begin{corollary} \label{cor:gef_consistency}
Assume a \GeF{} whose \GeDT{}s are learned under assumptions of Theorem~\ref{theo:consistency}. 
Then the \GeF{} of \GeDT{} predictors $p(Y \cbar \X_o)$, for any $\X_o \subseteq \X$, is Bayes consistent.
\end{corollary}

\section{Related Work}

Among the many variations of DTs and RFs that have been proposed in the last decades, the closest to our work are those that, similarly to \GeDT{}s and \GeF{}s, extend DT leaves with ``non-trivial'' models. Notable examples are DTs where the leaves are modelled by linear and logistic regressors \cite{Frank1998,Landwehr2005,Quinlan1992}, kernel density estimators (KDEs) \cite{Loh2009,Smyth1995}, linear discriminant models \cite{Gama2004,Kim2003}, KNN classifiers \cite{Buttrey2002,Loh2009}, and Naive-Bayes classifiers (NBCs) \cite{Kohavi1996}. Nonetheless, all these previous works focus primarily on improving the classification accuracy or smoothing probability estimates but do not model the full joint distribution, like in this work. Even extensions by Smyth et al.~\cite{Smyth1995} and Kohavi~\cite{Kohavi1996}, which include generative models (KDEs and NBCs, respectively) do not exploit their generative properties. To the best of our knowledge, GeFs are the first DT framework that effectively model and leverage the full joint distribution in a classification context. That is of practical significance as none of these earlier extensions of DTs offer a principled way to treat missing values or detect outliers. Here it is also worth mentioning the contemporary work of Khosravi et al. \cite{Khosravi2020} that proposes a similar probabilistic approach to handle missing data in DTs.

On the other side of the spectrum, DTs have also been extended to density estimators \cite{Gray2003,Rahman2014,Ram2011,Wu2014}. Among these, Density Estimation Trees (DETs) \cite{Ram2011}, Cutset Networks (CNets) \cite{Rahman2014}, and randomised ensembles thereof \cite{DiMauro2017}, are probably the closest to our work. These models are trained with a greedy tree-learning algorithm but minimise a modified loss function that matches their generative nature: joint entropy across all variables in CNets, mean integrated squared error in DETs. Notably, CNets, like GeFs, are Probabilistic Circuits, and hence also allow for tractable inference and marginalisation. They, however, have not been applied in a discriminative setting and are not backwards compatible with DTs and RFs. Moreover, \GeDT{}s (and \GeF{}s) can be seen as a family of models depending on the estimation at the leaves, making a clear parallel with what DTs (and RFs) offer.

Finally, one can also mimic the benefits of generative models in ensembles by learning predictors for all variables, as in MERCS \cite{Wolputte2018}. That is fundamentally different from our probabilistic approach and might entail prohibitively large numbers of predictors. Handling missing values, in the worst case, would require one predictor for each of the $2^{|\X|}$ missing patterns, and that is why MERCS relies on imputation methods when needed. Conversely, \GeDT{}s model a full joint distribution, thus being more compact and interpretable.

\section{Experiments}

\begin{figure}[t!]
    \begin{center}
        \input{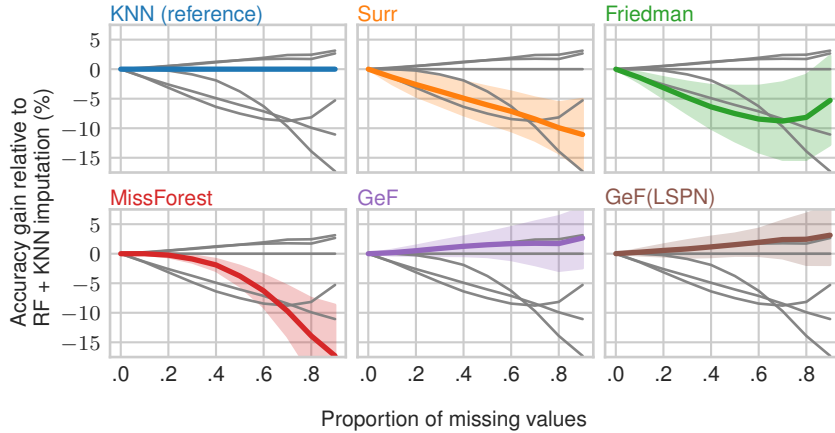}
    \end{center}
    \vspace{-.5cm}
    \caption{Average accuracy gain relative to RFs (100 trees) plus KNN imputation against proportion of missing values. The same plot is repeated six times, each time highlighting one method. The average as well as the confidence intervals (95\%) are computed across the 21 datasets of Table~\ref{tab:res}.}
    \label{fig:againstKNN}
\end{figure}

We run a series of classification tasks with incomplete data to compare our models against surrogate splits~\cite{Breiman1984,Therneau1997}, Friedman's method~\cite{Friedman1977,Quinlan1987b}, and mean (mode), KNN ($k=7$) and MissForest~\cite{Stekhoven2012} imputation. In particular, we experiment with two variants of \GeF{}s: one with fully-factorised leaves, which we denote simply \GeF{}, and another with leaves learned via LearnSPN \cite{Gens2013}, which we call \GeF(LearnSPN).We use a transformation of \GeF{}s into a clever PC that prunes unnecessary sub-trees \cite{Correia2019}, speeding up computations and achieving time complexity comparable to the original DTs and RFs (see Appendix B). In all experiments, \GeF, \GeF(LearnSPN) and the RF share the exact same structure (partition over the feature space) and are composed of 100 trees; including more trees has been shown to yield only marginal gains in most cases \cite{Probst2018}. In \GeF(LearnSPN), we run LearnSPN only for leaves with more than 30 samples, defaulting to a fully factorised model in smaller leaves.

We compare the accuracy of the methods in a selection of datasets from the OpenML-CC18 benchmark\footnote{\url{https://www.openml.org/s/99/data}} \cite{OpenML2013} and the wine-quality dataset \cite{moro2011}. Table~\ref{tab:res} presents results for 30\% of missing values at test time (different percentages are shown in Appendix C), with 95\% confidence intervals across 10 repetitions of 5-fold cross-validation. \GeF{} models outperform other methods in almost all datasets, validating that the joint distributions at the leaves provide enough information for computing the marginalisation in Eq.~\eqref{eq:marginalized_classifier}. We also note that increasing the expressive power of the models at the leaves seems worthwhile, as \GeF(LSPN) outperforms the vanilla \GeF{} in about half of the datasets. Similar conclusions are supported by Figure~\ref{fig:againstKNN}, where we plot the average gain in accuracy relative to RF + KNN imputation at different proportions of missing values. While earlier built-in methods, Friedman's and surrogate splits, perform poorly (justifying the popularity of imputation techniques for RFs), \GeF{}s are on average more than 3\% more accurate than KNN imputation. For the sake of space, a thorough exposition of these experiments is deferred to Appendix C, where we fully describe the experimental procedure, show different percentages of missing data and include results with PCs learned via \emph{class-selective} LearnSPN \cite{Correia2019}, as baseline for a standard generative model.

The reviewers suggested a direct comparison against CNets \cite{Rahman2014} since, like \GeF{}s, they are based on DTs and encode a proper joint distribution over all the variables. However, while we acknowledge the value of such comparison, the implementations to which we had access either did not support missing data or were too slow to yield reliable experimental results with ensembles of similar size, in the short time available to revise the paper. Also, it is worth noticing that 
CNets are currently not available for mixed variables, which prevents their application to most datasets in Table~\ref{tab:res}.

\begin{filecontents*}{Results/res30.csv}
Dataset,n,cat,cont,nc,maj,Mean,Surr,Friedman,KNN,MissForest,GeF,GeF(LSPN),Mci,Sci,Fci,Kci,MFci,GeFci,GeF(LSPN)ci,lower
dresses,500,12,0,2,58.0,58.18,45.48,55.8,56.62,55.68,57.12,57.14,0.85,1.57,1.21,1.57,0.95,1.11,1.07,57.33
wdbc,569,0,30,2,62.74,94.92,94.96,94.96,95.59,94.92,95.64,96.26,0.58,0.36,0.35,0.41,0.36,0.47,0.42,95.84
diabetes,768,0,8,2,65.1,71.67,72.97,73.35,72.4,72.46,73.93,73.83,0.84,0.73,0.7,0.75,0.92,0.63,0.72,73.3
vehicle,846,0,18,4,25.77,63.27,71.61,67.12,71.77,70.69,72.39,72.77,1.05,0.92,0.79,1.01,1.33,1.13,1.27,71.5
vowel,990,2,10,11,9.09,64.51,78.79,70.81,85.62,81.85,89.25,89.59,0.83,0.86,1.45,0.66,1.1,0.77,0.91,88.68
credit-g,1000,13,7,2,70.0,73.01,71.97,72.42,73.06,73.03,73.81,73.97,0.64,0.32,0.31,0.65,0.78,0.38,0.36,73.61
mice,1080,0,77,8,13.89,84.91,95.84,91.01,97.7,96.08,98.38,99.06,1.03,0.44,0.77,0.5,0.52,0.32,0.15,98.91
authent.,1372,0,4,2,55.54,84.47,88.65,87.13,91.98,90.74,90.33,89.66,0.79,0.74,0.69,0.55,0.37,0.65,0.64,91.43
cmc,1473,8,1,3,42.7,47.67,48.7,49.8,48.38,48.28,49.96,50.08,0.86,0.79,0.38,0.58,0.9,1.03,1.05,49.03
segment,2310,2,15,7,14.29,78.34,93.32,84.14,94.25,93.21,93.42,93.41,0.68,0.27,0.69,0.32,0.41,0.2,0.34,93.93
dna,3186,180,0,3,51.91,83.91,90.53,77.23,89.31,90.76,87.42,82.99,0.43,0.31,0.36,0.33,0.34,0.19,0.25,90.42
splice,3190,60,0,3,51.88,84.65,86.09,84.76,89.06,86.18,91.1,85.69,0.28,0.46,0.65,0.53,0.26,0.5,0.53,90.6
krvskp,3196,36,0,2,52.22,83.58,73.62,82.81,86.58,86.24,88.35,88.65,0.64,0.92,0.73,0.43,0.56,0.39,0.32,88.33
robot,5456,0,24,4,40.41,89.39,91.74,84.73,92.74,91.72,92.97,94.67,0.28,0.23,0.57,0.25,0.37,0.28,0.2,94.47
texture,5500,0,40,11,9.09,84.24,95.31,89.85,97.13,95.4,95.93,97.12,0.42,0.15,0.34,0.15,0.17,0.15,0.13,96.98
wine,6497,0,11,2,80.34,83.2,84.49,82.45,85.73,85.95,85.22,85.85,0.15,0.17,0.1,0.26,0.12,0.18,0.19,85.83
gesture,9873,0,32,5,29.88,55.41,58.37,52.86,61.62,61.48,58.65,60.2,0.21,0.15,0.27,0.22,0.26,0.18,0.21,61.4
phishing,11055,30,0,2,55.69,88.02,81.52,88.98,92.06,91.18,92.99,93.3,0.19,0.5,0.17,0.13,0.18,0.08,0.06,93.24
bank,41188,11,9,2,88.73,90.09,90.42,90.3,90.64,90.4,90.79,90.77,0.11,0.18,0.15,0.14,0.19,0.21,0.21,90.58
jungle,44819,6,0,3,51.46,66.89,63.45,71.91,66.25,65.67,72.4,72.3,0.4,0.24,0.12,0.15,0.2,0.12,0.11,72.28
electricity,45312,1,7,2,57.55,73.24,79.79,77.47,80.55,81.21,82.23,82.64,0.19,0.09,0.1,0.11,0.1,0.12,0.11,82.53
\end{filecontents*}
\DTLloaddb[]{res30}{Results/res30.csv}
\DTLforeach{res30}{\l=lower}
{
    \def\theMax{0} 
    \DTLforeachkeyinrow{\thisValue}{                
        \ifthenelse{\dtlcol>6}{                     
            \DTLmax{\theMax}{\theMax}{\thisValue}   
        }{}
    } 

    \DTLforeachkeyinrow{\thisValue}{                
        \ifthenelse{\dtlcol>6}{                     
            \ifthenelse{\DTLisFPgteq{\thisValue}{\l}}{
                \DTLreplaceentryforrow{\dtlkey}{\textbf{\thisValue}}
            }{}
            \ifthenelse{\DTLisieq{\thisValue}{\theMax}}{
                \DTLreplaceentryforrow{\dtlkey}{\underline{\textbf{\theMax}}}
            }{}
        }{}
    }     
}

\begin{table}[t!]
\fontsize{8}{8}\selectfont
\setlength{\tabcolsep}{2.25pt}
    \caption{Accuracy at 30\% of missing values at test time with 95\% confidence intervals. The best performing model is underlined, whereas all models within its confidence interval appear in bold.}
    \label{tab:res}
    \begin{tabular}{c c | c c c c c | c c}
    \toprule Dataset & n & Surrogate & Friedman & Mean & KNN & MissForest & \GeF & \GeF(LSPN)
    \DTLforeach*{res30}{\d=Dataset, \n=n, \cat=cat, \cont=cont, \nc=nc, \maj=maj,
                        \surr=Surr, \fried=Friedman, \mean=Mean, \knn=KNN, \mf=MissForest,
                        \gef=GeF, \gefl=GeF(LSPN),
                        \sci=Sci, \frci=Fci, \mci=Mci, \kci=Kci, \mfci=MFci,
                        \gefci=GeFci, \geflci=GeF(LSPN)ci}
    {
        \DTLiffirstrow{\\ \midrule}{\\}
        
        \d & \n &
        \surr~$\scriptstyle\pm$~\pgfmathprintnumber[res]\sci &
        \fried~$\scriptstyle\pm$~\pgfmathprintnumber[res]\frci &
        \mean~$\scriptstyle\pm$~\pgfmathprintnumber[res]\mci &
        \knn~$\scriptstyle\pm$~\pgfmathprintnumber[res]\kci &
        \mf~$\scriptstyle\pm$~\pgfmathprintnumber[res]\mfci &
        \gef~$\scriptstyle\pm$~\pgfmathprintnumber[res]\gefci &
        \gefl~$\scriptstyle\pm$~\pgfmathprintnumber[res]\geflci 
    }
    \\\bottomrule
    \end{tabular}
\end{table}

Another advantage of generative models is the ability of using the likelihood over the explanatory variables to detect outliers. \GeF{}s are still an ensemble of generative \GeDT{}s and thus do not encode a single full joint distribution. However, we can extend \GeF{}s to model a single joint by considering a uniform mixture of \GeDT{}s (using a sum node), instead of an ensemble of the conditional distributions of each \GeDT. In this case, the model represents the joint $p(\X,Y)=\ntrees^{-1}\sum_{j=1}^{\ntrees} p_j(\X,Y)$, where each $p_j$ comes from a different \GeDT. This model is named \Gp{} and achieves similar but slightly inferior performance than \GeF{}s in classification with missing data (still clearly superior to KNN imputation). This does not come as a surprise: the benefits of a fully generative models often comes at the cost of a (small) drop in classification accuracy (results in Appendix C).

\begin{figure}[ht!]
    \begin{center}
        \input{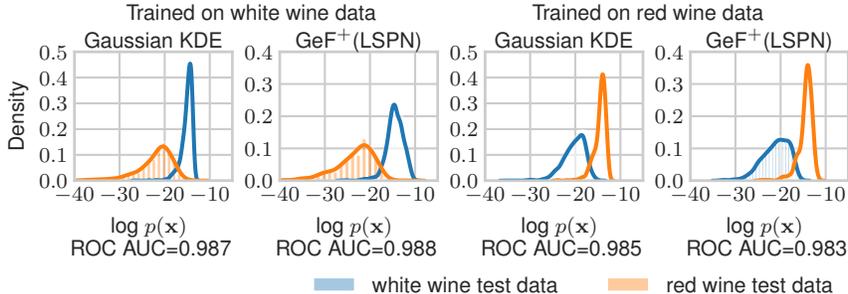}
    \end{center}
    \vspace{-.75cm}
    \caption{Normalised histograms of $\log p(\x)$ for samples from two different wine datasets.}
    \label{fig:fx_wine}
\end{figure}

We illustrate how to detect outliers with \GeF{}s by applying a \Gp(LSPN) to the the wine dataset \cite{Cortez2009} with a variant of transfer testing \cite{Bradshaw2017}. We learn two different \Gp(LSPN) models, each with only one type of wine data (red or white), to predict whether a wine has a score of 6 or higher. We then compute the log-density of unseen data (70/30 train test split) for the two wine types with both models. As we see in the histograms of Figure~\ref{fig:fx_wine}, the marginal distribution over explanatory variables does provide a strong signal to identify out-of-domain instances. In comparison to a Gaussian Kernel Density Estimator (KDE), \Gp(LSPN) achieved similar results even though its structure has been fit in a discriminative way.

We repeat a similar experiment with images, where we use the MNIST dataset \cite{Lecun1998} to fit a Gaussian KDE, a Random Forest and its corresponding \Gp. We  then evaluate these models on different digit datasets, namely Semeion \cite{Dua2019} and SVHN \cite{Netzer2011} (converted to grayscale and 784 pixels), to see whether they can identify out-of-distribution samples. We also use the entropy over the class variable as a baseline, since this is a signal that is easily computed on a standard Random Forest. Again, \Gp successfully identified out-of-domain samples, outperforming the two other methods and even encoding slightly different distributions for SVHN and Semeion digits.
Note that in both experiments we also compare the methods in terms of the area under the receiver operating characteristic curve (AUC ROC), which we computed using the log-density (or entropy) as a signal for a binary classifier that discriminates between in- and out-of-domain samples.

\begin{figure}[ht!]
    \begin{center}
        \input{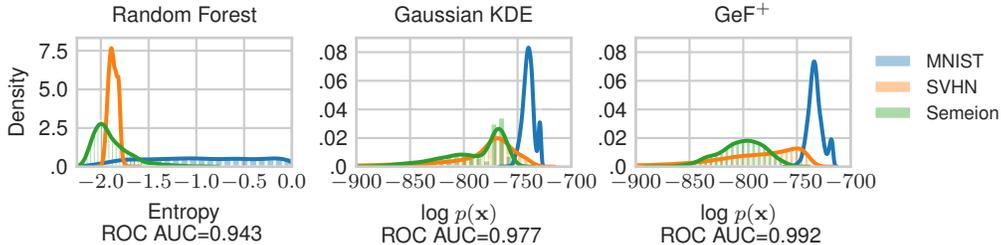}
    \end{center}
    \vspace{-.5cm}
    \caption{Normalised histograms of $\log p(\x)$ for samples from three different image datasets.}
    \label{fig:fx_images}
\end{figure}

\section{Conclusion}

By establishing a connection between Decision Trees (DTs) and Probabilistic Circuits (PCs), we have upgraded DTs to a full joint model over both inputs and outputs, yielding their generative counterparts, called \GeDT{}s. The fact that \GeDT{}s, and their ensemble version \GeF{}s, are ``backwards compatible'' to DTs and RFs, while offering benefits like consistent classification under missing inputs and outlier detection, makes it easy to adopt them in everyday practice. Missing data and outliers, however, are just the beginning. We believe that many of the current challenges in machine learning, like explainability, interpretability, and (adversarial) robustness are but symptoms of an overemphasis on purely discriminative methods in the past decades, and that hybrid generative approaches---like the one in this paper---will contribute significantly towards mastering these current challenges.

\newpage
\section*{Broader Impact}
This work establishes a connection between two sub-fields in machine learning, namely decision trees/random forests and probabilistic circuits.
Since there was very restricted communication between these two research communities, a fruitful cross-fertilisation of ideas, theory and algorithms between these research domains can be expected.
This represents a highly positive impact on fundamental machine learning and artificial intelligence research.

Decision trees and random forests are a de facto standard classification and regression tools in daily applied machine learning and data science.
Being---so far---purely discriminative models, they struggle with two problems which are key concerns in this work: missing data and outlier detection.
Since the improvements suggested in this paper can be incorporated in existing decision tree algorithms with very minor changes, our results have a potentially dramatic and immediate impact on a central and widely used machine learning and data science tool.

Since our work is elementary machine learning research, its ethical consequences are hard to assess.
However, the main ethical and societal impact of our work is the extension of a standard prediction tool, increasing its application domain and pertinence, and thus amplifying existing ethical considerations of data-driven and automatic prediction.

\begin{ack}
The authors thank the reviewers for their useful insights and suggestions. During part of the three years prior to the submission of this work, the authors were affiliated with the following institutions besides TU Eindhoven: Alvaro Correia was a full-time employee at Accenture and Ita\'u-Unibanco, and affiliated with Utrecht University; Cassio de Campos was affiliated with Queen's University Belfast and Utrecht University; Robert Peharz was affiliated with the University of Cambridge.
\end{ack}

\bibliography{GeFs.bib}
\bibliographystyle{abbrv}

\newpage
\begin{appendices}
\input{proofs.tex}

\newpage
\input{complexity.tex}

\newpage
\input{exp-missing.tex}

\end{appendices}

\end{document}

%% file: tree1.tex
\begin{center}

\definecolor{blu}{RGB}{199, 206, 234}
\definecolor{gre}{RGB}{181, 234, 215}
\definecolor{re}{RGB}{255, 154, 162}
\definecolor{ore}{RGB}{255, 218, 193}
\definecolor{lgr}{RGB}{226, 240, 203}
\definecolor{mel}{RGB}{255, 183, 178}

\begin{tikzpicture}[
    roundnode/.style={minimum size=7.5mm,
                      inner sep=0}, 
    cross/.style={minimum size=5mm, fill=lgr,
        path picture={
            \draw[black] (path picture bounding box.south east) -- (path picture bounding box.north west) (path picture bounding box.south west) -- (path picture bounding box.north east);
        }
    },
    sum/.style={minimum size=5mm, fill=blu,
        path picture={
            \draw[black] (path picture bounding box.south) -- (path picture bounding box.north) (path picture bounding box.west) -- (path picture bounding box.east);
        }
    },
    gauss/.style={minimum size=5mm, fill=ore,
        path picture={
            \draw[black] plot[domain=-.15:.15] ({\x},{exp(-200*\x*\x -2.)});
        }
    },
    dt/.style={minimum size=5mm, fill=white
    },
    line/.style={
      draw,thick,
      -latex',
      shorten >=2pt
    },
    cloud/.style={
      draw=red,
      thick,
      ellipse,
      fill=red!20,
      minimum height=1em
    }
]
    
    \node[draw, circle, dt] (D1) at (4, 3) { };
    \node[draw, circle, dt] (D2) at (3, 2) { };
    \node[draw, circle, dt] (D3) at (5, 2) { };
    \node[draw, circle, dt] (D4) at (2, 1) { };
    \node[draw, circle, dt] (D5) at (4, 1) { };
    
    \node (ind) at (5.225, 2.8) {$X_2>.5$};
    \node (ind) at (4.0, 1.775) {$X_1=1$};
    \node (ind) at (5.25, 1.5) {$(20,0)$};
    \node (ind) at (4.0, 0.5) {$(10,30)$};
    \node (ind) at (2.0, 0.5) {$(0,40)$};

    \draw[line width=0.3mm, <-, auto] (D2.45) to node[black]{$X_2\leq.5$} (D1.225);
    \draw[line width=0.3mm, <-] (D3.135) -- (D1.315);
    \draw[line width=0.3mm, <-, auto] (D4.45) to node[black]{$X_1=0$} (D2.225);
    \draw[line width=0.3mm, <-] (D5.135) --  (D2.315);

    \node[draw, circle, sum] (Dz1) at (4+5, 3) { };
    \node[draw, circle, sum] (Dz2) at (3+5, 2) { };
    \node[draw, circle, gauss] (Dz3) at (5+5, 2) { };
    \node[draw, circle, gauss] (Dz4) at (2+5, 1) { };
    \node[draw, circle, gauss] (Dz5) at (4+5, 1) { };
    
    \node (ind) at (4.7+5, 2.8) {$.2$};
    \node (ind) at (3.7+5, 1.775) {$.5$};
    \node (ind) at (6.4+5, 1.5) {$p_3(X_1,X_2,Y)\cdot \indfu{X_2 > 0.5}$};
    \node (ind) at (4.4+5, 0.5) {$p_2(X_1,X_2,Y)\cdot$};
    \node (ind) at (4.4+5, 0.1) {$\indfu{X_1=1}\cdot \indfu{X_2\leq 0.5}$};
    \node (ind) at (1.7+5, 0.5) {$p_1(X_1,X_2,Y)\cdot$};
    \node (ind) at (1.7+5, 0.1) {$\indfu{X_1=0}\cdot \indfu{X_2\leq 0.5}$};

    \draw[line width=0.3mm, <-, auto] (Dz2.45) to node[black]{$.8$} (Dz1.225);
    \draw[line width=0.3mm, <-] (Dz3.135) -- (Dz1.315);
    \draw[line width=0.3mm, <-, auto] (Dz4.45) to node[black]{$.5$} (Dz2.225);
    \draw[line width=0.3mm, <-] (Dz5.135) --  (Dz2.315);

\end{tikzpicture}
\end{center}

%% file: proofs.tex
\section{Theoretical Results}
\begin{proposition}
A \GeDT{} is \emph{deterministic}.
\end{proposition}
\begin{proof}
Consider any sum node $\node$ in a \GeDT{} and assume, for simplicity, that it has two children $\nodec'$ and $\nodec''$.
Node $\node$ is associated with a partition $\{\xxs_{\nodec'}, \xxs_{\nodec''}\}$ of $\xxs_{\node}$.
Any leaf $\leaf$ which is a descendant of $\nodec'$, respectively $\nodec''$, must have a support which is a subset of $\xxs_{\nodec'}$, respectively $\xxs_{\nodec''}$.
Assume that $\nodec'(\x) > 0$ for certain $\x$, implying $\x \in \xxs_{\nodec'}$ and thus $\x \notin \xxs_{\nodec''}$.
Therefore, $\nodec''(\x) = 0$, since $\x$ is not in the support of any leaf below $\nodec''$.
The same argument holds for the reverse case and straightforwardly extends to arbitrarily many sum nodes. Thus $\node$ is deterministic.
\end{proof}

\begin{reptheorem}{theo:backwards_compatible}
Let $f$ be a DT classifier and $p(Y \cbar \x)$ be a corresponding \GeDT{} classifier, where each leaf in \GeDT{} is class-factorised, i.e.~of the form $p(Y)p(\X)$, and where $p(Y)$ has been estimated in the maximum-likelihood sense.
Then $f(\x) = p(Y \cbar \x)$, provided that $p(\x) > 0$.
\end{reptheorem}
\begin{proof}
Recall that the leaves in the \GeDT{} are in one-to-one correspondence with the leaf cells $\lcells$ of the DT, and that the support of any leaf is given by its corresponding $\lcell \in \lcells$.
Let $\node_{\x}$ be the unique leaf in the \GeDT{} whose cell is $\lcell(\x)$.
Since \GeDT{} is a tree-shaped PC containing only sum nodes, its joint distributions is either $p_{\node_\x}(\x, y)$---if \GeDT{} consists only of $\node_{\x}$---or can be written as 
\begin{equation}   \label{eq:GeDT_sum}
p(\x,y) = \sum_{\nodec \in \ch(\node)} w_{\node,\nodec} \nodec(\x),
\end{equation}
where $\node$ is the root node.
Since the \GeDT{} is deterministic, it has at most one non-zero child.
From $p(\x) > 0$ it follows that the \GeDT{} has \emph{exactly} one non-zero child, say $\nodec'$, and \eqref{eq:GeDT_sum} can be written as $p(\x,y) = \w_{\node,\nodec'} \nodec'(\x, y)$.
Now, since $\nodec'(\x,y)$ is also a tree-shape PC containing only sums, it follows by induction that $p(\x, y) = \left( \prod_{(\node, \nodec) \in \pat} w_{\node,\nodec} \right) p_{\node_\x}(\x,y)$, where $\pat$ is the unique path from root to $\node_\x$ following only non-zero nodes, and $\w_{\node,\nodec}$ are the sum-weights of edges $(\node,\nodec)$ in $\pat$.
Since each leaf is class-factorised, we have $p_{\node_\x}(\x, y) = p_{\node_\x}(\x) p_{\node_\x}(y)$, and
$
\left( \prod_{(\node, \nodec) \in \pat} w_{\node,\nodec} \right) p_{\node_\x}(\x) p_{\node_\x}(y)
\propto
p(y \cbar \x) 
= p_{\node_\x}(y)
= f^{\lcell(\x)}(\x)
= f(\x)$,
since each $f^\lcell(\x)$ is---like $p_{\node_\x}$---learned by the class proportions of samples falling in $\lcell$.
\end{proof}

\begin{reptheorem}{theo:consistency}
Let $\prd^*$ be an unknown data generating distribution with density $p^*(\X,Y)$, and let $\mathcal{D}_n$ be a dataset drawn i.i.d.~from $\prd^*$.
Let $\graph$ be a DT learned with a DT learning algorithm, using axis-aligned splits.
Let $\lcells^n$ be the (rectangular) leaf cells produced by the learning algorithm.
Assume it holds that 
i) $\lim_{n \rightarrow \infty}\nicefrac{|\lcells^n| \log(n)}{n} \rightarrow 0$ and ii) $\prd^*(\{\x ~|~ \mathrm{diam}(\lcell^n_\x) > \gamma\}) \rightarrow 0$ almost surely for all $\gamma > 0$, where $\mathrm{diam}(\lcell)$ is the diameter of cell $\lcell$.
Let $\graph'$ be the \GeDT{} corresponding to $\graph$, obtained via Algorithm~1, where for each leaf $v$, $p_v$ is of the form $p_v(Y) p_v(\X)$, with $p_v(\X)$ uniform on $\lcell_v$ and $p_v(Y)$ the maximum likelihood Categorical (fractions of class values of samples in $\lcell_v$).
Then the \GeDT{} distribution is $l_1$-consistent, i.e.~$\sum_y \int | p(\x,y) - p^*(\x,y) | \mathrm{d}\x \rightarrow 0$, almost surely.
\end{reptheorem}

Before proving Theorem~\ref{theo:consistency} we need to introduce some background.
This theorem extends consistency results for collections of partitions of the state space $\xxs$, as discussed by Lugosi and Nobel~\cite{Lugosi1996}.
A central notion is the \emph{growth function} of such partitions.
\begin{definition}[Growth function \cite{Lugosi1996}]   \label{def:growth_function}
Let $\xxs$ be some set and $\lcellsfamily$ be a collection of finite partitions of $\xxs$.
Let $\bm{\xi} = \{\xi_1, \dots, \xi_n\}$ be a set of points from $\xxs$.
Let $\Delta(\lcellsfamily, \bm{\xi})$ be the number of distinct partitions induced by $\lcellsfamily$, that is the size of set $\{\{\bm{\xi} \cap \lcell ~|~ \lcell \in \lcells\} ~|~ \lcells \in \lcellsfamily\}$.
The growth function is defined as $\Delta^*(\lcellsfamily) = \sup_{\bm{\xi}} \Delta(\lcellsfamily, \bm{\xi})$, where the $\sup$ ranges over all sets of $n$ points from $\xxs$.
\end{definition}
Note that the growth function $\Delta^*$ is defined akin to the \emph{dichotomic growth function}, as introduced by Vapnik and Chervonenkis and well known in statistical learning theory \cite{Vapnik1998}.
In particular, we derive the following bound of $\Delta^*$.
\begin{proposition}   \label{prop:growth_function_bound}
Let $\xxs$ be some set and $\bm{\mathcal{C}}$ be any collection of subsets of $\xxs$.
Let $\Phi(\bm{\mathcal{C}}, \bm{\xi})$ be the shatter coefficient of point set $\bm{\xi}$ and $\Phi^*(\bm{\mathcal{C}}) = \sup_{\bm{\xi}} \Phi(\bm{\mathcal{C}}, \bm{\xi})$ be the dichotomic growth function \cite{Vapnik1998}.
Let $\lcellsfamily$ be a collection of finite partitions of $\xxs$, as in Definition~\ref{def:growth_function}, where the maximal partition size is $J := \sup_{\lcells \in \lcellsfamily} |\lcells|$.
If $\bm{\mathcal{C}} = \{ \lcell ~|~  \lcell \in \lcells, \lcells \in \lcellsfamily \}$ then \begin{equation}   \label{eq:growthfunction_inequality}
\Delta(\lcellsfamily, \bm{\xi}) \leq \Phi(\bm{\mathcal{C}}, \bm{\xi})^J,
\end{equation}
and moreover $\Delta^*(\lcellsfamily) \leq \Phi^*(\bm{\mathcal{C}})^J$.
\end{proposition}
\begin{proof}
Let the point set $\bm{\xi}$ be fixed.
Any partition $\{\bm{\xi} \cap \lcell ~|~ \lcell \in \lcells\}$, for some $\lcells \in \lcellsfamily$, can be written as $\{\bm{\xi} \cap \lcell_1, \dots, \bm{\xi} \cap \lcell_J\}$ for some $\lcell_1, \dots, \lcell_J \in \bm{\mathcal{C}}$, since $\bm{\mathcal{C}}$ contains all cells which appear in $\lcellsfamily$.
Thus, $\Delta(\lcellsfamily, \bm{\xi}) \leq |\{\{\bm{\xi} \cap \lcell_1, \dots, \bm{\xi} \cap \lcell_J\} ~|~ \lcell_1, \dots, \lcell_J \in \bm{\mathcal{C}} \}|$.
Note that the number of partitions of this form is bounded by
\begin{equation}   \label{eq:number_partitions_bound}
|\{\{\bm{\xi} \cap \lcell_1, \dots, \bm{\xi} \cap \lcell_J\} ~|~ \lcell_1, \dots, \lcell_J \in \bm{\mathcal{C}} \}| \leq 
\bigtimes_{j=1}^J |\{\bm{\xi} \cap \lcell_j ~|~ \lcell_j \in \bm{\mathcal{C}} \}|.
\end{equation}
The right hand side of \eqref{eq:number_partitions_bound} is $\Phi(\bm{\mathcal{C}}, \bm{\xi})^J$, and thus \eqref{eq:growthfunction_inequality} follows. 
$\Delta^*(\lcellsfamily) \leq \Phi^*(\bm{\mathcal{C}})^J$ follows from applying $\sup_{\bm{\xi}}$ on both sides of \eqref{eq:growthfunction_inequality}.
\end{proof}

In our case, we study partitions $\lcells$ induced by a DT, each of which divides $\xxs$ into a set of hyper-rectangles.\footnote{Here, we assume for simplicity that all variables are continuous. Including discrete variables with finitely many states can be done by applying similar arguments to each of the finitely many joint states.}
Hence, we consider the collection of partitions $\lcellsfamily$ containing all possible partitions whose sets are hyper-rectangles.
We are now ready to prove Theorem~\ref{theo:consistency}.

\begin{proof}
Let $\lcellsfamily^n$ be the collection of all DT partitions which can be generated for sample size $n$, i.e.~$\lcells^n \in \lcellsfamily^n$.
By Proposition~\ref{prop:growth_function_bound}, we know that $\Delta^*(\lcellsfamily^n) \leq \Phi^*(\bm{\mathcal{C}})^{|\lcells^n|}$, where $\bm{\mathcal{C}}$ is the collection of all sub-rectangles in $\xxs$.
The VC dimension \cite{Vapnik1998} of $\bm{\mathcal{C}}$ is known to be  $2|\X|$, and consequently, by Sauer's lemma, $\Delta^*(\lcellsfamily) \leq \Phi^*(\bm{\mathcal{C}})^{|\lcells^n|} \leq C n^{2 |\lcells^n| |\X|}$, where $C$ is a constant depending only on $|\X|$.
Therefore, if condition i) holds ($\lim_{n \rightarrow \infty}\nicefrac{|\lcells^n| \log(n)}{n} \rightarrow 0$) it follows that $\frac{\log \Delta^*}{n} \rightarrow 0$.
Thus, together with condition ii) all conditions of Theorems 1 and 2 in \cite{Lugosi1996} hold.

Since the \GeDT{} is deterministic, its distribution can be written as
\begin{equation}   \label{eq:proof_consistent_GeDT}
p(\x, y) = \left( \prod_{(\node, \nodec) \in \pat} w_{\node,\nodec} \right) p_{\node_\x}(\x,y),    
\end{equation}
where $\node_\x$ is the unique non-zero leaf in the \GeDT{}, $\pat$ is the unique path from the root to $\node_\x$ following only non-zero nodes, and $\w_{\node,\nodec}$ are the sum-weights of edges $(\node,\nodec)$ in $\pat$ (see also proof of Theorem~\ref{theo:backwards_compatible}).

It is easy to see that $\prod_{(\node, \nodec) \in \pat} w_{\node,\nodec} = \hat{\mathbb{P}}(\lcell_\x)$, where $\hat{\mathbb{P}}$ is the empirical distribution of $\data_n$, i.e.~the fraction of data points falling in $\lcell_\x$ (see Algorithm~1 in the main paper).
The distribution computed by each leaf $\node$ is, by assumption, $p_{\node}(\x,y) = p_{\node}(y) \frac{1}{\mathrm{vol}(\lcell_\x)}$, where $\mathrm{vol}(\lcell)$ is the volume (Lebesgue measure) of $\lcell$.
Thus, we can write \eqref{eq:proof_consistent_GeDT} as 
\begin{equation}   \label{eq:proof_consistent_GeDT_2}
p(\x, y) = p_{\node}(y) \hat{\mathbb{P}}(\lcell_\x) \frac{1}{\mathrm{vol}(\lcell_\x)}.
\end{equation}

By Theorem 1 in \cite{Lugosi1996}, $\hat{\mathbb{P}}(\lcell_\x) \frac{1}{\mathrm{vol}(\lcell_\x)}$ converges to $p^*(\x)$, while 
by Theorem 2 in \cite{Lugosi1996}, $p_\node(y)$ converges to $p^*(y \cbar \x)$, both in $l1$-sense.
Clearly both factors, $\hat{\mathbb{P}}(\lcell_\x) \frac{1}{\mathrm{vol}(\lcell_\x)}$ and $p_\node(y)$, have bounded $l1$-norm.
Thus, their product converges to $p^*(y \cbar \x) p^*(\x) = p^*(y,\x)$, which concludes the proof.
\end{proof}

\begin{repcorollary}{cor:gedt_consistency}
Under assumptions of Theorem~\ref{theo:consistency}, any \GeDT{} predictor $p(Y \cbar \X_o)$, for $\X_o \subseteq \X$ is Bayes consistent.
\end{repcorollary}

\begin{proof}
Since $p(y,\x)$ converges almost surely to $p^*(y,\x)$ in $l1$-sense, it gives rise to the Bayes optimal classifier $\arg\max_y p^*(y, \x)$.
Consider any $X_i \in \X$.
The marginal distribution, $X_i$ marginalised out, is $\int p(y,\x_{\lnot i}, x_i) \mathrm{d}x_i$.
Since 
\begin{align}
\int | p(y,\x_{\lnot i}) - p^*(y,\x_{\lnot i}) | \mathrm{d}\x_{\lnot i}
& =
\int \left | \int p(y,\x_{\lnot i}, x_i) - p^*(y,\x_{\lnot i}, x_i) \mathrm{d}x_i \right | \mathrm{d}\x_{\lnot i} \\
& \leq 
\int |p(y,\x) - p^*(y,\x) | d\x,
\end{align}
also the marginal converges in $l1$-sense to the true $p^*(y,\x_{\lnot i})$.
By repeating the argument, every sub-marginal converges, and thus gives rise to the corresponding Bayes optimal classifier.
\end{proof}

\begin{repcorollary} {cor:gef_consistency}
Assume a \GeF{} whose \GeDT{}s are learned under assumptions of Theorem~\ref{theo:consistency}. 
Then the \GeF{} of \GeDT{} predictors $p(Y \cbar \X_o)$, for any $\X_o \subseteq \X$, is Bayes consistent.
\end{repcorollary}

\begin{proof}
This follows directly from Proposition 1 in \cite{Biau2008}, whereby if a sequence of classifiers is Bayes-consistent, then the classifier obtained by averaging them is also consistent.
\end{proof}

%% file: complexity.tex
\section{Time Complexity}
Let $n$ be the total number of samples and $m$ the total number of features.
Regarding the learning algorithm, a Random Forest and its corresponding PC only differ in the distributions at leaves, which use a partition of the data. Therefore, assuming a tree is grown as in \cite{Breiman2001} with $\lceil m/c\rceil$ features considered at each split ($c$ a positive natural), structure learning in both models has worst-case asymptotic complexity of $\asymO(m r~n\log n)$, where $r\in\asymO(n)$ is the number of internal nodes in the obtained tree \cite{Louppe2014}. For \GeDT{}s, however, there is the additional cost of learning a distribution at each leaf. If $q(m)$ is the worst-case cost of the leaf learner for a constant amount of data, then the overall time complexity (for learning all leaves) is $\asymO(r~q(m))$.

Nonetheless, if the leaf learner is such that $q(m)\leq\asymO(m n \log n)$, then the complexity is dominated by the structure learning and Random Forests and \GeF{}s have the same worst-case asymptotic complexity of $\asymO(n_t~(m r~n\log n + r q(m)))\leq\asymO(n_t~m r~n\log n)$, where $n_t$ is the number of trees in the model. Note that $q(m)\leq\asymO(m n \log n)$ holds for many learning algorithms when only a small number of training samples fall in each leaf---namely, LearnSPN and fully-factorised leaves---provided the reasonable assumption that $m$ is $\asymO(n)$.

 \begin{figure*}[h!]
     \centering
     \input{tree3.tex}
     \caption{Illustration of pulling indicators up to speed up computations (in the example, $\X$ and $Y$ factorise at leaves). On top, the original decision tree (DT) is shown. Below, both models represent the probabilistic circuit for the original DT and encode the very same distribution, even if the one in the right-hand side is not decomposable.}
     \label{fig:spn2spn}
 \end{figure*}
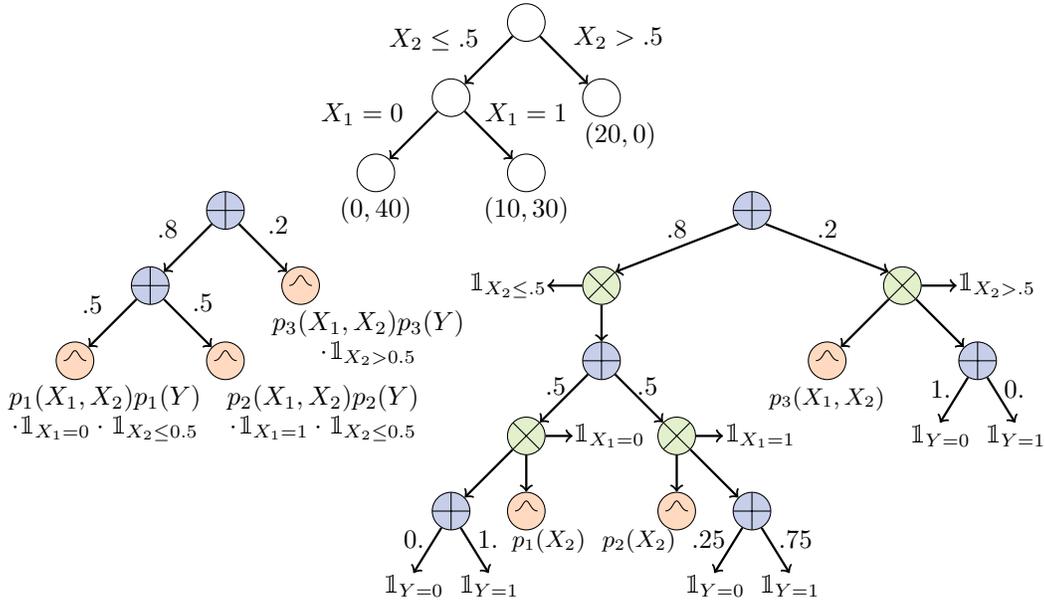

To perform inference for a complete test sample, \GeDT{}s require traversing the whole structure once (hence time $\asymO(r)$), while DTs have a worst-case of $\asymO(d)$, where $d$ is the height of the tree. However, we can bring the complexity of \GeDT{}s down to $\asymO(d)$ by placing the indicators that define the decisions of the internal nodes of the DT near the corresponding internal nodes of the \GeDT{}. This requires augmenting \GeDT{}s with product nodes, one for each internal sum node. Every new product node has two children: a sum node and an indicator mimicking the decision tree split, that is, the indicator only evaluates to one if that path in the tree is active. Figure~\ref{fig:spn2spn} illustrates the idea using the running example of the main paper, where the densities are as follows
\begin{align*}
    p_1(X_1,X_2,Y) &= p_1(X_1,X_2)  (0\cdot \indf{Y=0} + 1\cdot \indf{Y=1}),\\
    p_2(X_1,X_2,Y) &= p_2(X_1,X_2) (0.25\cdot \indf{Y=0} + 0.75\cdot \indf{Y=1}),\\
    p_3(X_1,X_2,Y) &= p_3(X_1,X_2) (1\cdot \indf{Y=0} + 0\cdot \indf{Y=1}).
\end{align*} 
This idea does not change results, since it is the same as bringing the common indicators that appeared in the leaves of a sub-tree up towards the root of that sub-tree using the distributive property of multiplication (for the PC enthusiast, the lack of decomposability is tackled by the determinism of the indicators). By evaluating indicators as soon as possible in a top-down recursive computation, we can avoid computing all sub-trees for which a zero is returned to a product node. With this type of computational graph, \GeF{}s and RFs have a similar inference procedure. Predicting the class of an instance amounts to traversing each tree and evaluating the corresponding leaf, and thus the inference complexity is $\asymO(n_t d)$.

For incomplete data, however, \GeDT{}s need to reach every active leaf (just as Friedman's method). Assuming the number of missing values in each instance is bounded by a constant, \GeF{}s still take time $\asymO(n_t d)$, being faster than  Random Forests with KNN imputation, which in the worst case take time $\asymO(n_t d + nm)$. For large (non-constant) percentages of missing values, \GeF{}s can be as slow as $\asymO(n_t r)$ (as it may need to reach all leaves). In this case of large numbers of missing values per instance, \GeF{}s are faster than Random Forests with KNN imputation if $d\approx r$ but slower if $d \ll r$.

%% file: tree3.tex
\begin{center}

\definecolor{blu}{RGB}{199, 206, 234}
\definecolor{gre}{RGB}{181, 234, 215}
\definecolor{re}{RGB}{255, 154, 162}
\definecolor{ore}{RGB}{255, 218, 193}
\definecolor{lgr}{RGB}{226, 240, 203}
\definecolor{mel}{RGB}{255, 183, 178}

\begin{tikzpicture}[
    roundnode/.style={minimum size=7.5mm,
                      inner sep=0}, 
    cross/.style={minimum size=5mm, fill=lgr,
        path picture={
            \draw[black] (path picture bounding box.south east) -- (path picture bounding box.north west) (path picture bounding box.south west) -- (path picture bounding box.north east);
        }
    },
    sum/.style={minimum size=5mm, fill=blu,
        path picture={
            \draw[black] (path picture bounding box.south) -- (path picture bounding box.north) (path picture bounding box.west) -- (path picture bounding box.east);
        }
    },
    gauss/.style={minimum size=5mm, fill=ore,
        path picture={
            \draw[black] plot[domain=-.15:.15] ({\x},{exp(-200*\x*\x -2.)});
        }
    },
    dt/.style={minimum size=5mm, fill=white
    },
    line/.style={
      draw,thick,
      -latex',
      shorten >=2pt
    },
    cloud/.style={
      draw=red,
      thick,
      ellipse,
      fill=red!20,
      minimum height=1em
    }
]

    \node[draw, circle, dt] (D1) at (4+3, 3+2.5) { };
    \node[draw, circle, dt] (D2) at (3+3, 2+2.5) { };
    \node[draw, circle, dt] (D3) at (5+3, 2+2.5) { };
    \node[draw, circle, dt] (D4) at (2+3, 1+2.5) { };
    \node[draw, circle, dt] (D5) at (4+3, 1+2.5) { };
    
    \node (ind) at (5.225+3, 2.8+2.5) {$X_2>.5$};
    \node (ind) at (4.0+3, 1.775+2.5) {$X_1=1$};
    \node (ind) at (5.25+3, 1.5+2.5) {$(20,0)$};
    \node (ind) at (4.0+3, 0.5+2.5) {$(10,30)$};
    \node (ind) at (2.0+3, 0.5+2.5) {$(0,40)$};
    
     \draw[line width=0.3mm, <-, auto] (D2.45) to node[black]{$X_2\leq.5$} (D1.225);
    \draw[line width=0.3mm, <-] (D3.135) -- (D1.315);
    \draw[line width=0.3mm, <-, auto] (D4.45) to node[black]{$X_1=0$} (D2.225);
    \draw[line width=0.3mm, <-] (D5.135) --  (D2.315);
    
    \node[draw, circle, sum] (Dz1) at (4-1, 3) { };
    \node[draw, circle, sum] (Dz2) at (3-1, 2) { };
    \node[draw, circle, gauss] (Dz3) at (5-1, 2) { };
    \node[draw, circle, gauss] (Dz4) at (2-1, 1) { };
    \node[draw, circle, gauss] (Dz5) at (4-1, 1) { };
    
    \node (ind) at (4.7-1, 2.8) {$.2$};
    \node (ind) at (3.7-1, 1.775) {$.5$};
    \node (ind) at (5.4-0.5, 1.5) {$p_3(X_1,X_2)p_3(Y)$};
    \node (ind) at (5.4-0.5, 1.1) {$\cdot \indfu{X_2 > 0.5}$};
    \node (ind) at (4.2+0.1, 0.5) {$p_2(X_1,X_2)p_2(Y)$};
    \node (ind) at (4.2+0.1, 0.1) {$\cdot \indfu{X_1=1}\cdot \indfu{X_2\leq 0.5}$};
    \node (ind) at (1.8-0.4, 0.5) {$p_1(X_1,X_2)p_1(Y)$};
    \node (ind) at (1.8-0.4, 0.1) {$\cdot \indfu{X_1=0}\cdot \indfu{X_2\leq 0.5}$};

    \draw[line width=0.3mm, <-, auto] (Dz2.45) to node[black]{$.8$} (Dz1.225);
    \draw[line width=0.3mm, <-] (Dz3.135) -- (Dz1.315);
    \draw[line width=0.3mm, <-, auto] (Dz4.45) to node[black]{$.5$} (Dz2.225);
    \draw[line width=0.3mm, <-] (Dz5.135) --  (Dz2.315);

    \node[draw, circle, sum] (S1) at (10, 3) { };

    \node[draw, circle, cross] (P1) at (8, 2) { };
    \node[draw, circle, sum] (S2) at (8, 1) { };
    \node[roundnode](B1) at (6.75, 2) {$\indfu{X_2\leq.5}$};
    
    \node[draw, circle, cross] (P2) at (12, 2) { };
    \node[draw, circle, sum] (S3) at (13, 1) { };
    \node[roundnode](C0'') at (12.5, 0) {$\indfu{Y=0}$};
    \node[roundnode](C1'') at (13.5, 0) {$\indfu{Y=1}$};
    \node[roundnode] (B2) at (13.25, 2) {$\indfu{X_2>.5}$};
    \node[draw, circle, gauss] (S3') at (11, 1) { };
    \node (b) at (11, 0.5) {\footnotesize $p_3(X_1,X_2)$};

    \node[draw, circle, cross] (P3) at (7, 0) { };
    \node[roundnode] (A0) at (8.1, 0) {$\indfu{X_1=0}$};
    \node[draw, circle, gauss] (G2) at (7, -1) {};
    \node (b) at (7.3, -1.4) {\footnotesize $p_1(X_2)$};
    \node[draw, circle, sum] (S4) at (6, -1) { };
    \node[roundnode](C0) at (5.5, -2) {$\indfu{Y=0}$};
    \node[roundnode](C1) at (6.5, -2) {$\indfu{Y=1}$};
    
    \node[draw, circle, cross] (P4) at (9, 0) { };
    \node[roundnode] (A1) at (10.1, 0) {$\indfu{X_1=1}$};
    \node[draw, circle, gauss] (G3) at (9, -1) { };
    \node (b) at (8.5, -1.4) {\footnotesize $p_2(X_2)$};
    \node[draw, circle, sum] (S5) at (10, -1) { };
    \node[roundnode](C0') at (9.5, -2) {$\indfu{Y=0}$};
    \node[roundnode](C1') at (10.5, -2) {$\indfu{Y=1}$};
    
    \draw[line width=0.3mm, <-, above] (P1.45) to node[black]{$.8$} (S1.225);
    \draw[line width=0.3mm, <-, above] (P2.135) to node[black]{$.2$} (S1.315);
    
    \draw[line width=0.3mm, <-, left, pos=0.75] (P3.45) to node[black]{$.5$} (S2.225);
    \draw[line width=0.3mm, <-, right, pos=0.75] (P4.135) to node[black]{$.5$} (S2.315);

    \draw[line width=0.3mm, <-] (S2.90) -- (P1.270);
    \draw[line width=0.3mm, <-] (B1.0) -- (P1.180);
    \draw[line width=0.3mm, <-] (S3.135) -- (P2.315);
    \draw[line width=0.3mm, <-] (B2.180) -- (P2.0);
    \draw[line width=0.3mm, <-] (A0.180) -- (P3.0);
    \draw[line width=0.3mm, <-] (A1.180) -- (P4.0);
    \draw[line width=0.3mm, <-] (G2.90) -- (P3.270);
    \draw[line width=0.3mm, <-] (G3.90) -- (P4.270);
    \draw[line width=0.3mm, <-] (S4.45) -- (P3.225);
    \draw[line width=0.3mm, <-] (S5.135) -- (P4.315);
    \draw[line width=0.3mm, <-] (S3'.45) -- (P2.225);
    
    \draw[line width=0.3mm, <-, left, pos=0.75] (C0''.center)+(0pt, 5pt) to node[black]{$1.$} (S3.240);
    \draw[line width=0.3mm, <-, right,pos=0.75] (C1''.center)+(0pt, 5pt) to node[black]{$0.$} (S3.300);
    
    \draw[line width=0.3mm, <-, left, pos=0.75] (C0'.center)+(0pt, 5pt) to node[black]{$.25$} (S5.240);
    \draw[line width=0.3mm, <-, right, pos=0.75] (C1'.center)+(0pt, 5pt) to node[black]{$.75$} (S5.300);
    
    \draw[line width=0.3mm, <-, left, pos=0.75] (C0.center)+(0pt, 5pt) to node[black]{$0.$} (S4.240);
    \draw[line width=0.3mm, <-, right, pos=0.75] (C1.center)+(0pt, 5pt) to node[black]{$1.$} (S4.300);
    
\end{tikzpicture}
\end{center}

%% file: exp-missing.tex
\section{Missing Values Experimental Results}
All 21 datasets are listed here in alphabetical order. For each of them, we report (both in tabular and graphic formats) the accuracy values at different percentages of missing data at test time, with 95\% confidence intervals. These confidence intervals are computed across 10 repetitions of 5-fold cross validation with different random seeds. The datasets were obtained directly from the OpenML-CC18 benchmark web-page \footnote{\url{https://www.openml.org/s/99/data}} \cite{OpenML2013}, and the only pre-processing step was standardising continuous features (mean $\mu=0$ and standard deviation $\sigma=1$) and mapping categorical features to $\{0, \ldots, K_i-1\}$. The datasets as well as the source code are provided with the supp. material.

We also present a few relevant details of each dataset.
\begin{itemize}
    \item[] n: number of samples.
    \item[] m$_0$: number of categorical variables.
    \item[] m$_1$: number of numerical variables.
    \item[] $|\mathcal Y|$: number of classes.
    \item[] \%Maj: percentage of the majority class.
\end{itemize}

For the sake of completeness, we briefly discuss each of the methods and their implementations. The source code is all in Python 3 and all experiments were run in a single laptop with a modern CPU.

\paragraph{Random Forest implementation}
In all experiments, the structure of all models is kept the same, that is, they are all derived from the same Random Forest and thus share the same partition of the feature space. For every dataset, the Random Forests were composed of 100 ``deep'' trees, that is, the only stop criterion is the impurity of the class variable, possibly leading to many leaves with a single sample. Each tree is learned on a bootstrap sample of the same size of the training dataset, and each split only evaluates $\sqrt{m}$ variables, with $m$ the total number of features.We use the Gini impurity measure as the criterion to select the best split in the decision-tree learning and rank surrogate splits according to how well they predict the best split, as in \cite{Therneau1997}. The trees are all binary, with splits on categorical variables defined by two subsets of the possible instantiations. That is somewhat different from other implementations, where the splits are either ``full", yielding one child per category, or given by a threshold, which implicitly assumes categorical variables are ordinal.

\paragraph{``Built-in'' Methods}
These are methods for treating missing values that do not require external models, and hence are ``built-in'' into the decision tree structure. In fact, they consist of slight modifications to the inference procedure.
\begin{itemize}
    \item[] \textbf{Surrogate splits} \cite{Breiman1984,Therneau1997}:
    During training, once the best split is defined, one ranks alternative splits on the number of instances that they send to the same branch as the best split. At test time, if the split variable is not observed, one tries the surrogate splits in order (starting with that which most resembles the best split). If none of the surrogate split variables is available, the instance is sent to the branch with the highest number of data points at training time. Surrogate splits have two notable drawbacks: (i) their performance is heavily dependent on the correlation between variables; (ii) they require storing every possible split to be guaranteed to work for all missing-value configurations, which is rather computational intensive, especially for large ensembles.
    \item [] \textbf{Friedman method} \cite{Friedman1977,Quinlan1987b}: Whenever a split variable is not observed, one follows both branches of the tree. That means any instance with missing value is mapped to multiple leaves, and the final prediction is given by the majority class across the sum of the counts of all these leaves. If $C^{\lcell}(j)$ gives the number of training instances of class $j$ in cell $\lcell$, we can write Friedman's methods as
    \begin{equation*}
        \begin{split}
            f(\x) = \argmax_{j \in \{1, \ldots, K\}} \sum_{\lcell \in \lcells} \indf{\x \in \lcell}C^{\lcell}(j), \\
            C^{\lcell}(j) = \sum_{i=1}^n \indf{\x_i \in \lcell}\indf{y_i = j},
        \end{split}
    \end{equation*}
    
    where $i$ runs through the $n$ training instances $(\x_i, y_i)$, and $j$ runs through the $K$ possible classes. Note that Friedman's method can be seen as a simplified version of \GeF{}s where the density over explanatory variables is constant and the same in every leaf.
\end{itemize}

\paragraph{Imputation methods} It is not surprising that most of the work on handling missing data in decision trees and random forests rely on data imputation \cite{Saar-Tsechansky2007}. That is, another or multiple other models are used to predict the missing values before feeding the data to the tree-based classifier. In the experiments we compare two different types of imputation methods: 
\begin{itemize}
    \item[] \textbf{Mean} Missing values are imputed with the mean for continuous variables or the most frequent observation for categorical variables.
    \item[] \textbf{KNN} Similar to the simple method above but the means or most frequent values are taken over the $K$-nearest neighbours. We use a standard K-nearest neighbour implementation from \texttt{scikit-learn} \cite{scikit-learn} with K=7. However, the distance function is updated to better accommodate mixed data types. Following, Huang et al. \cite{Huang1997}, we define the distance measure as
    $$d(\x_a, \x_b) = \gamma \sum_{i=0}^{m_0} w_i \delta(\x_a[i], \x_b[i]) + \sum_{i=m_0}^{m_1} w_i \sqrt{(\x_a[i] - \x_b[i])^2},$$
    where $\gamma$ is a parameter representing the relative importance of categorical and numerical features, $w_i$ is the weight of feature $i$, and, without loss of generality, we assume features are ordered so that the first m$_0$ variables are categorical. The $\delta$ function is simply the Hamming distance: $\delta(\x_a[i], \x_b[i])=1$ if $\x_a[i] \neq \x_b[i]$, and $\delta(\x_a[i], \x_b[i])=0$ otherwise. As we have no reason to favour any feature or feature type, we set both $\gamma$ and every $w_i$ to one.
    \item[] \textbf{MissForest} \cite{Stekhoven2012}: For each variable $X_i \in \X$, one learns a Random Forest (classifier/regressor) that is used to predict unobserved values of $X_i$ given the other variables $\X \setminus X_i$. As more than one variable might be unobserved, MissForest starts by imputing missing values with the mean (or mode) and then iteratively updates its initial guess using the Random Forest predictors. The original MissForest algorithm proposed in \cite{Stekhoven2012} also updates the Random Forest predictors at every iteration. However, in our experiments that would allow MissForest to exploit test data information, which could compromise the results. Therefore, we fit the Random Forest predictors in the training data only and keep them fixed at test time. Note that the algorithm remains iterative, since the imputed values are still fed to the predictors in the next iteration. We use a standard Python implementation of MissForests from \texttt{missingpy}---adapted to accommodate the changes mentioned above---which relies on the \texttt{scikit-learn} implementation of Random Forests.
\end{itemize}

\paragraph{Vanilla GeFs}
What we call \emph{vanilla} \GeF{}, or simply \GeF{}, is a model where the distribution at the leaves is given by a fully factorised model, that is, for each leaf $v$, $p_{v}(\x, y) = p_{v}(x_1)p_{v}(x_2) \ldots p_{v}(x_m)p_{v}(y).$
This is probably the simplest model that one can fit at the leaves and is clearly \emph{class-factorised}. Therefore, vanilla \GeF{}s preserve full backward-compatibility with the original RF, yielding the exact same prediction function for complete data.

\paragraph{\GeF{} with LearnSPN}
For GeF(LearnSPN) and GeF$^+$(LearnSPN), the LearnSPN algorithm \cite{Gens2013} is run only at leaves with more than 30 samples, and smaller leaves are modelled by a fully factorised model as in vanilla GeFs. That saves computational time with little performance impact, as the model derived from LearnSPN with few samples would be similarly simplistic. We run the LearnSPN algorithm as follows: sum nodes split the samples via K-means clustering with K=2, and product nodes split the variables with an independence threshold of 0.001 (pair of variables for which the independence test yields a p-value lower than the threshold are considered independent). We do not force independence between the class $Y$ and input variables $\bm{X}$ in LearnSPN, which explains why, in contrast to GeF, GeF(LearnSPN) does not necessarily yield the same predictions as the original Random Forest.

\paragraph{LearnSPN}
Similarly, we also learn a Probabilistic Circuit by applying the LearnSPN algorithm \cite{Gens2013} to the entire dataset. The hyperparameters for this experiment are the same as for \GeF{}s with LearnSPN, but we use a variant of LearnSPN that yields class-selective PCs, which have been shown to outperform standard LearnSPN in classification tasks \cite{Correia2019}.

\newpage
\subsection[(Banknote) Authentication]{(Banknote) Authentication \cite{Dua2019}}

\begin{table}[h!]
    \centering
    \begin{tabular}{c c c c c}
        \multicolumn{5}{c}{Dataset details}\\ \\
        n & m$_0$ & m$_1$ & $|\mathcal Y|$ & \%Maj\\\hline
        1372 & 0 & 4 & 2 & 55.54
    \end{tabular}
\end{table}

\maketable{Results/authent.csv}{tab:authent}

\begin{figure}[h!]
    \begin{center}
        \scalebox{0.9}{\input{Figures/authent.pgf}}
    \end{center}
    \caption{\mycap}
\end{figure}

\newpage
\subsection[Bank Marketing]{Bank Marketing \cite{moro2011}}

\begin{table}[h!]
    \centering
    \begin{tabular}{c c c c c}
        \multicolumn{5}{c}{Dataset details}\\ \\
        n & m$_0$ & m$_1$ & $|\mathcal Y|$ & \%Maj\\\hline
        41188 &	11 & 9 & 2	& 88.73 
    \end{tabular}
\end{table}

\maketable{Results/bank.csv}{tab:bank}

\begin{figure}[h!]
    \begin{center}
        \scalebox{0.9}{\input{Figures/bank.pgf}}
    \end{center}
    \caption{\mycap}
\end{figure}

\newpage
\subsection[Breast Cancer (WDBC)]{Breast Cancer (WDBC) \footnote{This breast cancer domain was obtained from the University Medical Centre, Institute of Oncology, Ljubljana, Yugoslavia. Thanks go to M. Zwitter and M. Soklic for providing the data.}}

\begin{table}[h!]
    \centering
    \begin{tabular}{c c c c c}
        \multicolumn{5}{c}{Dataset details}\\ \\
        n & m$_0$ & m$_1$ & $|\mathcal Y|$ & \%Maj\\\hline
        569 & 0 & 30 & 2 & 62.74
    \end{tabular}
\end{table}

\maketable{Results/breast.csv}{tab:breast}

\begin{figure}[h!]
    \begin{center}
        \scalebox{.8}{\input{Figures/breast.pgf}}
    \end{center}
    \caption{\mycap}
\end{figure}

\newpage
\subsection[Contraceptive Method Choice (CMC)]{Contraceptive Method Choice (CMC) \cite{Dua2019}}

\begin{table}[h!]
    \centering
    \begin{tabular}{c c c c c}
        \multicolumn{5}{c}{Dataset details}\\ \\
        n & m$_0$ & m$_1$ & $|\mathcal Y|$ & \%Maj\\\hline
        1473 &	8 & 1 & 3 & 42.7
    \end{tabular}
\end{table}

\maketable{Results/cmc.csv}{tab:cmc}

\begin{figure}[h!]
    \begin{center}
        \scalebox{0.9}{\input{Figures/cmc.pgf}}
    \end{center}
    \caption{\mycap}
\end{figure}

\newpage
\subsection[Credit-g]{Credit-g \cite{Dua2019}}

\begin{table}[h!]
    \centering
    \begin{tabular}{c c c c c}
        \multicolumn{5}{c}{Dataset details}\\ \\
        n & m$_0$ & m$_1$ & $|\mathcal Y|$ & \%Maj\\\hline
        1000 & 13 &	7 &	2 &	70.0
    \end{tabular}
\end{table}

\maketable{Results/german.csv}{tab:german}

\begin{figure}[h!]
    \begin{center}
        \scalebox{0.9}{\input{Figures/german.pgf}}
    \end{center}
    \caption{\mycap}
\end{figure}

\newpage
\subsection[Diabetes]{Diabetes \cite{Dua2019}}

\begin{table}[h!]
    \centering
    \begin{tabular}{c c c c c}
        \multicolumn{5}{c}{Dataset details}\\ \\
        n & m$_0$ & m$_1$ & $|\mathcal Y|$ & \%Maj\\\hline
        768 & 0 & 8 & 2 & 65.1
    \end{tabular}
\end{table}

\maketable{Results/diabetes.csv}{tab:diabetes}

\begin{figure}[h!]
    \begin{center}
        \scalebox{0.9}{\input{Figures/diabetes.pgf}}
    \end{center}
    \caption{\mycap}
\end{figure}

\newpage
\subsection[DNA (Primate splice-junction gene sequences)]{DNA (Primate splice-junction gene sequences) \cite{Dua2019}}

\begin{table}[h!]
    \centering
    \begin{tabular}{c c c c c}
        \multicolumn{5}{c}{Dataset details}\\ \\
        n & m$_0$ & m$_1$ & $|\mathcal Y|$ & \%Maj\\\hline
        3186 & 180 & 0 & 3 & 51.91
    \end{tabular}
\end{table}
This is the same dataset as Splice, but here the categorical variables were one-hot encoded.
\maketable{Results/dna.csv}{tab:dna}

\begin{figure}[h!]
    \begin{center}
        \scalebox{0.8}{\input{Figures/dna.pgf}}
    \end{center}
    \caption{\mycap}
\end{figure}

\newpage
\subsection[Dresses-sales]{Dresses-sales \cite{Dua2019}}

\begin{table}[h!]
    \centering
    \begin{tabular}{c c c c c}
        \multicolumn{5}{c}{Dataset details}\\ \\
        n & m$_0$ & m$_1$ & $|\mathcal Y|$ & \%Maj\\\hline
        500	& 12 & 0 & 2 & 58.0
    \end{tabular}
\end{table}

\maketable{Results/dresses.csv}{tab:dresses}

\begin{figure}[h!]
    \begin{center}
        \scalebox{0.9}{\input{Figures/dresses.pgf}}
    \end{center}
    \caption{\mycap}
\end{figure}

\newpage
\subsection[Electricity]{Electricity \cite{Gama2004}}

\begin{table}[h!]
    \centering
    \begin{tabular}{c c c c c}
        \multicolumn{5}{c}{Dataset details}\\ \\
        n & m$_0$ & m$_1$ & $|\mathcal Y|$ & \%Maj\\\hline
        45312 &	1 & 7 &	2 &	57.55
    \end{tabular}
\end{table}

\maketable{Results/electricity.csv}{tab:electricity}

\begin{figure}[h!]
    \begin{center}
        \scalebox{0.9}{\input{Figures/electricity.pgf}}
    \end{center}
    \caption{\mycap}
\end{figure}

\newpage
\subsection[Gesture Phase Segmentation]{Gesture Phase Segmentation \cite{Madeo2013}}

\begin{table}[h!]
    \centering
    \begin{tabular}{c c c c c}
        \multicolumn{5}{c}{Dataset details}\\ \\
        n & m$_0$ & m$_1$ & $|\mathcal Y|$ & \%Maj\\\hline
        9873 & 0 & 32 &	5 &	29.88
    \end{tabular}
\end{table}

\maketable{Results/gesture.csv}{tab:gesture}

\begin{figure}[h!]
    \begin{center}
        \scalebox{0.9}{\input{Figures/gesture.pgf}}
    \end{center}
    \caption{\mycap}
\end{figure}

\newpage
\subsection[Jungle Chess]{Jungle Chess \cite{OpenML2013}}

\begin{table}[h!]
    \centering
    \begin{tabular}{c c c c c}
        \multicolumn{5}{c}{Dataset details}\\ \\
        n & m$_0$ & m$_1$ & $|\mathcal Y|$ & \%Maj\\\hline
        44819 &	6 &	0 &	3 &	51.46
    \end{tabular}
\end{table}

\maketable{Results/jungle.csv}{tab:jungle}

\begin{figure}[h!]
    \begin{center}
        \scalebox{0.9}{\input{Figures/jungle.pgf}}
    \end{center}
    \caption{\mycap}
\end{figure}

\newpage
\subsection[King-Rook vs. King-Pawn (kr-vs-kp)]{King-Rook vs. King-Pawn (kr-vs-kp) \cite{Dua2019}}

\begin{table}[h!]
    \centering
    \begin{tabular}{c c c c c}
        \multicolumn{5}{c}{Dataset details}\\ \\
        n & m$_0$ & m$_1$ & $|\mathcal Y|$ & \%Maj\\\hline
        3196 & 36 &	0 &	2 &	52.22
    \end{tabular}
\end{table}

\maketable{Results/krvskp.csv}{tab:krvskp}

\begin{figure}[h!]
    \begin{center}
        \scalebox{0.9}{\input{Figures/krvskp.pgf}}
    \end{center}
    \caption{\mycap}
\end{figure}

\newpage
\subsection[Mice Protein]{Mice Protein \cite{Higuera2015}}

\begin{table}[h!]
    \centering
    \begin{tabular}{c c c c c}
        \multicolumn{5}{c}{Dataset details}\\ \\
        n & m$_0$ & m$_1$ & $|\mathcal Y|$ & \%Maj\\\hline
        1080 & 0 & 77 &	8 &	13.89
    \end{tabular}
\end{table}

\maketable{Results/mice.csv}{tab:mice}

\begin{figure}[h!]
    \begin{center}
        \scalebox{0.9}{\input{Figures/mice.pgf}}
    \end{center}
    \caption{\mycap}
\end{figure}

\newpage
\subsection[Phishing Websites]{Phishing Websites \cite{Dua2019}}

\begin{table}[h!]
    \centering
    \begin{tabular}{c c c c c}
        \multicolumn{5}{c}{Dataset details}\\ \\
        n & m$_0$ & m$_1$ & $|\mathcal Y|$ & \%Maj\\\hline
        11055 & 30 & 0 & 2 & 55.69
    \end{tabular}
\end{table}

\maketable{Results/phishing.csv}{tab:phishing}

\begin{figure}[h!]
    \begin{center}
        \scalebox{0.9}{\input{Figures/phishing.pgf}}
    \end{center}
    \caption{\mycap}
\end{figure}

\newpage
\subsection[Robot (Wall-Following Robot Navigation)]{Robot (Wall-Following Robot Navigation) \cite{Dua2019}}

\begin{table}[h!]
    \centering
    \begin{tabular}{c c c c c}
        \multicolumn{5}{c}{Dataset details}\\ \\
        n & m$_0$ & m$_1$ & $|\mathcal Y|$ & \%Maj\\\hline
        5456 & 0 & 24 & 4 & 40.41
    \end{tabular}
\end{table}

\maketable{Results/robot.csv}{tab:robot}

\begin{figure}[h!]
    \begin{center}
        \scalebox{0.9}{\input{Figures/robot.pgf}}
    \end{center}
    \caption{\mycap}
\end{figure}

\newpage
\subsection[Segment]{Segment \cite{Dua2019}}

\begin{table}[h!]
    \centering
    \begin{tabular}{c c c c c}
        \multicolumn{5}{c}{Dataset details}\\ \\
        n & m$_0$ & m$_1$ & $|\mathcal Y|$ & \%Maj\\\hline
        2310 & 2 & 15 & 7 & 14.29
    \end{tabular}
\end{table}

\maketable{Results/segment.csv}{tab:segment}

\begin{figure}[h!]
    \begin{center}
        \scalebox{0.9}{\input{Figures/segment.pgf}}
    \end{center}
    \caption{\mycap}
\end{figure}

\newpage
\subsection[Splice (Primate splice-junction gene sequences)]{Splice (Primate splice-junction gene sequences) \cite{Dua2019}}

\begin{table}[h!]
    \centering
    \begin{tabular}{c c c c c}
        \multicolumn{5}{c}{Dataset details}\\ \\
        n & m$_0$ & m$_1$ & $|\mathcal Y|$ & \%Maj\\\hline
        3190 & 60 & 0 & 3 & 51.88
    \end{tabular}
\end{table}

\maketable{Results/splice.csv}{tab:splice}

\begin{figure}[h!]
    \begin{center}
        \scalebox{0.9}{\input{Figures/splice.pgf}}
    \end{center}
    \caption{\mycap}
\end{figure}

\newpage
\subsection[Texture]{Texture \footnote{This database was generated by the Lab. of Image Processing and Pattern Recognition (INPG-LTIRF) in the development of the Esprit project ELENA No. 6891 and the Esprit working group ATHOS No. 6620.}}

\begin{table}[h!]
    \centering
    \begin{tabular}{c c c c c}
        \multicolumn{5}{c}{Dataset details}\\ \\
        n & m$_0$ & m$_1$ & $|\mathcal Y|$ & \%Maj\\\hline
        5500 & 0 & 40 & 11 & 9.09
    \end{tabular}
\end{table}

\maketable{Results/texture.csv}{tab:texture}

\begin{figure}[h!]
    \begin{center}
        \scalebox{0.8}{\input{Figures/texture.pgf}}
    \end{center}
    \caption{\mycap}
\end{figure}

\newpage
\subsection[Vehicle]{Vehicle \cite{Siebert1987}}

\begin{table}[h!]
    \centering
    \begin{tabular}{c c c c c}
        \multicolumn{5}{c}{Dataset details}\\ \\
        n & m$_0$ & m$_1$ & $|\mathcal Y|$ & \%Maj\\\hline
        846 & 0	& 18 & 4 & 25.77
    \end{tabular}
\end{table}

\maketable{Results/vehicle.csv}{tab:vehicle}

\begin{figure}[h!]
    \begin{center}
        \scalebox{0.9}{\input{Figures/vehicle.pgf}}
    \end{center}
    \caption{\mycap}
\end{figure}

\newpage
\subsection[Vowel]{Vowel \cite{Dua2019}}

\begin{table}[h!]
    \centering
    \begin{tabular}{c c c c c}
        \multicolumn{5}{c}{Dataset details}\\ \\
        n & m$_0$ & m$_1$ & $|\mathcal Y|$ & \%Maj\\\hline
        990	& 2 & 10 & 11 & 9.09
    \end{tabular}
\end{table}

\maketable{Results/vowel.csv}{tab:vowel}

\begin{figure}[h!]
    \begin{center}
        \scalebox{0.9}{\input{Figures/vowel.pgf}}
    \end{center}
    \caption{\mycap}
\end{figure}

\newpage
\subsection[Wine Quality]{Wine Quality \cite{moro2011}}

\begin{table}[h!]
    \centering
    \begin{tabular}{c c c c c}
        \multicolumn{5}{c}{Dataset details}\\ \\
        n & m$_0$ & m$_1$ & $|\mathcal Y|$ & \%Maj\\\hline
        6497 & 0 & 11 & 2 & 80.34
    \end{tabular}
\end{table}

This dataset includes both red and white wine data. For classification purposes, the target variable was split into two classes: scores less or equal to 6, and scores greater than 6.

\maketable{Results/wine.csv}{tab:wine}

\begin{figure}[h!]
    \begin{center}
        \scalebox{0.8}{\input{Figures/wine.pgf}}
    \end{center}
    \caption{\mycap}
\end{figure}